\documentclass{article}

\usepackage{microtype}
\usepackage{graphicx}
\usepackage{subfigure}
\usepackage{booktabs} 
\usepackage[dvipsnames]{xcolor}

\usepackage[noend]{algorithmic}

\usepackage[accepted]{icml2024}

\usepackage[colorlinks=true,linkcolor=red,urlcolor=Cerulean,citecolor=Green]{hyperref}%

\usepackage{multirow}
\usepackage{amsmath}
\usepackage{amssymb}
\usepackage{mathtools}
\usepackage{amsthm}

\usepackage[capitalize,noabbrev]{cleveref}

\theoremstyle{plain}
\newtheorem{theorem}{Theorem}[section]
\newtheorem{proposition}[theorem]{Proposition}

\theoremstyle{definition}
\newtheorem{definition}[theorem]{Definition}

\theoremstyle{remark}

\newcommand{\model}{Efficient Linear-Complexity Search Algorithm over \textbf{E}dge-\textbf{i}nduced Sub\textbf{G}raphs}
\newcommand{\smodel}{EiG-Search}

\newcommand{\black}[1]{\textcolor{black}{#1}}

\usepackage[textsize=tiny]{todonotes}

\icmltitlerunning{EiG-Search: Generating Edge-Induced Subgraphs for GNN Explanation in Linear Time}

\begin{document}

\twocolumn[


\icmltitle{EiG-Search: Generating Edge-Induced Subgraphs \\for GNN Explanation in Linear Time}



\icmlsetsymbol{equal}{*}

\begin{icmlauthorlist}
\icmlauthor{Shengyao Lu}{ua}
\icmlauthor{Bang Liu}{udem}
\icmlauthor{Keith G. Mills}{ua}
\icmlauthor{Jiao He}{hua}
\icmlauthor{Di Niu}{ua}
\end{icmlauthorlist}

\icmlaffiliation{ua}{Department of Electrical and Computer Engineering, University of Alberta}
\icmlaffiliation{hua}{Kirin AI Algorithm \& Solution, Huawei}
\icmlaffiliation{udem}{DIRO, Universit{\'e} de Montr{\'e}al \& Mila, Canada CIFAR AI Chair}

\icmlcorrespondingauthor{Shengyao Lu}{shengyao@ualberta.ca}

\icmlkeywords{Machine Learning, Explainable AI, Interpretability, Explainability, Graph Neural Networks, ICML}

\vskip 0.3in
]



\printAffiliationsAndNotice{}  

\begin{abstract}
Understanding and explaining the predictions of Graph Neural Networks (GNNs), is crucial for enhancing their safety and trustworthiness. Subgraph-level explanations are gaining attention for their intuitive appeal. However, most existing subgraph-level explainers face efficiency challenges in explaining GNNs due to complex search processes. The key challenge is to find a balance between intuitiveness and efficiency while ensuring transparency. Additionally, these explainers usually induce subgraphs by nodes, which may introduce less-intuitive disconnected nodes in the subgraph-level explanations or omit many important subgraph structures. 
In this paper, we reveal that inducing subgraph explanations by edges is more comprehensive than other subgraph inducing techniques. We also emphasize the need of determining the subgraph explanation size for each data instance, as different data instances may involve different important substructures. Building upon these considerations, we introduce a training-free approach, named {\smodel}. We employ an efficient linear-time search algorithm over the edge-induced subgraphs, where the edges are ranked by an enhanced gradient-based importance.
We conduct extensive experiments on a total of seven datasets, demonstrating its superior performance and efficiency both quantitatively and qualitatively over the leading baselines. Our code is available at: \href{https://github.com/sluxsr/EiG-Search}{\textit{https://github.com/sluxsr/EiG-Search}}. 
\end{abstract}

\section{Introduction}
\label{sec:intro}

The explainability of Graph Neural Networks (GNNs) has become a crucial topic, driven by their ``black box'' nature and the demand for transparency in sensitive fields. While earlier works focus on generating node-level or edge-level explanations \cite{pope2019explainability, baldassarre2019explainability, shrikumar2017learning, vu2020pgm, huang2022graphlime, ying2019gnnexplainer, luo2020parameterized, schlichtkrull2020interpreting, zhang2021relex, lin2021generative}, there is growing attention on subgraph-level explanations~\cite{yuan2021explainability, shan2021reinforcement, feng2022degree, zhang2022gstarx, ye2023same, li2023dag, pereira2023distill}, since they are more intuitive and human-understandable. 

However, existing subgraph-level explainers often involve sophisticated processes to generate subgraph explanations, resulting in inefficiency and limiting their practical applications. For example, MotifExplainer~\cite{yu2022motifexplainer} relies on costly expert knowledge to first identify subgraphs before passing them to the explainer. As another example, SubgraphX~\cite{yuan2021explainability} searches for the subgraph explanations with the Shapley value serving as the scoring function. Although employing the Monte Carlo Tree Search algorithm, their method is still computationally demanding. Therefore, it remains a key challenge to balance intuitiveness and efficiency of GNN explainability. 
Moreover, most existing subgraph-level explainers induce the subgraphs by node groups, which may result in disconnected nodes in the explanations, reducing the intuitiveness. Inducing by nodes also leads to non-exhaustive enumeration over the possible subgraphs, posing the risk of omitting important subgraph structures. Futhermore, they usually pre-specify a universally fixed number or ratio for the explanation size. Nevertheless, given that different data samples may have varying explanation sizes, this setup makes the explanations less convincing and reliable. 

Another line of existing GNN explainability approaches rely on a second auxiliary black-box model~\cite{ying2019gnnexplainer, luo2020parameterized, vu2020pgm, bajaj2021robust, shan2021reinforcement, lin2021generative, huang2022graphlime, li2023dag, pereira2023distill}. While these methods provide high quality explanations, they can be inconsistent across different runs. As pointed out by~\citet{zhao2023towards}, these methods may introduce non-deterministic behaviors even for the same input graph since they require training an auxiliary or secondary model. A lack of consistency will compromise the faithfulness of the explanation as well. In view of this issue, other studies utilize gradients or gradient back-propagation to determine the critical graph components (i.e., nodes, edges, subgraphs)~\cite{pope2019explainability, baldassarre2019explainability, schnake2021higher}. While being white-boxes and training-free, these techniques suffer from the gradient saturation problem~\cite{shrikumar2017learning}, affecting their explaining performance. More discussion on related work can be found in \cref{app:related_work}.

In this paper, we point out via analysis that edge-induced subgraph explanations are more intuitive and exhaustive than subgraphs typically induced by nodes or by nodes and edges in the literature. Moreover, we show that the size of the best explaining subgraph 
can vary between graph samples, and thus prior methods that find subgraphs at a specified size (or sparsity) for all the samples in a dataset may not be optimal. Based on these insights, we propose an {\model} (\smodel), which is a \emph{training-free} and efficient search procedure to generate the best subgraph-level GNN explanation for a given graph instance in linear time complexity, while also automatically searching for the optimal subgraph size. 

Unlike many existing subgraph-level explainers that typically employ intricate heuristic search methods and generate explaining subgraphs at a predetermined size, our efficient method generates the optimal subgraph by evaluating a reduced search space of subgraphs induced by sorted edges. In particular, for each edge, {\smodel} first utilizes an edge importance approximation algorithm that calculates a linear gradient of the original graph representation from a baseline graph representation with respect to that edge. Then, we perform a search over candidate subgraphs induced by top-$k$ edges, exhausting all values of $k$ to obtain the subgraph that maximizes the overall explanation performance.

Furthermore, different from existing gradient-based interpretation, the linear gradient that we use to approximate edge scores avoids direct manipulation of gradients. Instead, it constructs latent lines connecting base points to the original data points in space. We compute the gradients of the latent lines to represent edge importance, which will not ``saturate''. We further distinguish this mechanism from Integrated Gradients (IG)~\cite{sundararajan2017axiomatic} through both a discussion of its design and empirical results. The findings indicate that our Linear Gradients outperform IG on graph-related tasks.

We compare our approach with a range of leading subgraph-level GNN explanation methods to demonstrate the faithfulness and efficiency of {\smodel}. Also, we evaluate the efficacy of individual components in our method, including the linear-time search and edge importance approximation by augmenting existing methods with these proposed components. 
The results clearly show that {\smodel} yields significantly superior subgraph explanations compared to existing methods, while being remarkably more efficient. 

\section{Preliminary}
\label{sec:prelim}

\paragraph{Notations.} Let $G=(V,E)$ denotes a graph with a node feature matrix $\mathbf{X} \in \mathbb{R}^{n\times d}$, where each row of $\mathbf{X}$ represents the node feature vector $\mathbf{x}_v$ for $v\in V$. $d$ is the dimension of node features and $n=|V|$ represents the number of nodes in $G$. The graph adjacency matrix is $\mathbf{A}\in \mathbb{R}^{n\times n}$. A graph neural network could be written as $\phi(\mathbf{A},\mathbf{X})\rightarrow \mathbf{Y}$, which maps a graph to a probability distribution over a set of classes denoted by $\mathbf{Y}$.

\paragraph{Graph neural networks.} GNNs~\cite{kipf2017semisupervised, xu2018powerful} use the graph structure, namely the adjacency matrix $\mathbf{A}$, and the node features $\mathbf{X}$ to learn node representations $h_v$ for each node $v \in V$ or a graph representation $h_G$ of $G$, and then perform node/graph classification tasks. 
At each layer, GNNs update the representation of a node by aggregating its neighboring node representations. 
The node representation with a $L$-layer GNN can capture the structural information within its $L$-hop neighborhood. Formally, the representation vector $h^{(k)}_v$ of each node $v$ at the $k$-th layer is:
\begin{equation}
\label{eq:akv}
    a^{(k)}_v = \mathrm{AGG}^{(k)}\left ( \left \{ h^{(k-1)}_u: u \in \mathcal{N}(v) \right \} \right ),
\end{equation}
\begin{equation}
\label{eq:hkv}
    h^{(k)}_v = \mathrm{COMB}^{(k)}\left (h^{(k-1)}_v, a^{(k)}_v \right ),
\end{equation} 
where $\mathcal{N}(v)$ is a set of nodes adjacent to $v$, AGG is an aggregation function, and COMB is a combining function. 

\paragraph{Problem setup.}
This paper focuses on generating intuitive \emph{subgraph explanations} for \emph{instance-level} GNN explainability. 
In an instance-level GNN explanation task, we are given a pre-trained GNN: $\phi(\mathbf{A},\mathbf{X})\rightarrow \mathbf{Y}$ and a corresponding dataset $\mathcal{G}$, where $G\in \mathcal{G}$. In our paper, we aim to highlight the outstanding subgraphs within $G$ that are important to the GNN predictions for each instance $G$ in $\mathcal{G}$. We assess the subraph explanations using two commonly used metrics~\cite{haoyuansurvey2022, GNNBook2022} the edge removal-based counterfactual metric $Fidelity^+$ and the completeness metric $Fidelity^-$. The definitions of these metrics can be found in  \cref{sec:size_subgraph_exp}.

\section{Investigating Subgraph-level Explanations via Inducing Technique and Size}
\label{sec:declarations}

\begin{figure*}[t]
\begin{center}
    \vskip 0.05in
    \centerline{\includegraphics[width=0.9\textwidth]{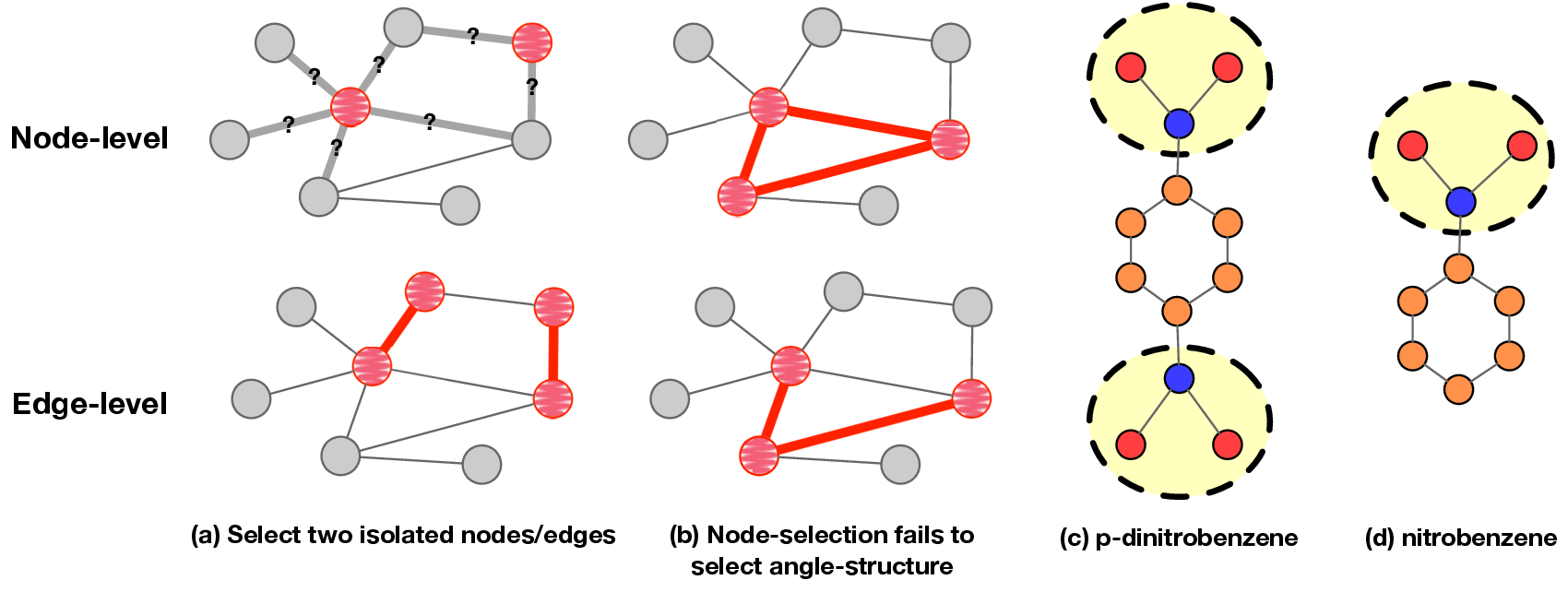}}
    \vskip -0.1in
    \caption{Illustration of subgraph explanations. (a): If nodes that are not directly neighboring each other are selected, determining the important subgraph structure becomes non-trivial. If edges are selected, the corresponding endpoints are naturally selected, which naturally gives a subgraph explanation. (b): Node-selection-based methods are not able to discover the angle-shape structure as an explanation, whereas edge-selection can be helpful. (c): The orange nodes stand for ``C", blue nodes stand for ``N", and red nodes stand for ``O". Picking a single subgraph for explanations cannot properly find the disconnected ``NO2'' groups as we highlighted. (d): The size of the critical subgraph is the size of highlighted ``NO2'', which is different from (c).}
    \label{fig:node_edge}
\end{center}
\vskip -0.2in
\end{figure*}

In this section, we provide a comprehensive study on the process of producing subgraph-level explanations via the perspectives of \emph{subgraph inducing technique} and \emph{explanation size}. In terms of the subgraph inducing technique, most existing subgraph-level GNN explanation approaches~\cite{yuan2021explainability, shan2021reinforcement, feng2022degree, zhang2022gstarx, pereira2023distill} utilize a node-induced technique to obtain the explanation subgraphs. However, we find that the edge-induced technique is better than the node-induced technique in providing intuitive subgraphs. On the other hand, most existing approaches rely on human experts to manually determine the appropriate size of subgraph explanations. For example, DEGREE~\cite{feng2022degree} restricts explanatory subgraphs to contain $q$ nodes, while RG-Explainer~\cite{shan2021reinforcement} confines node-induced subgraphs to contain $k$ edges. Here, $q$ and $k$ are hyperparameters representing the size of the presumed ``ground-truth'' subgraph explanations. While such presumptions may be effective on synthetic datasets, applying them to real-world tasks becomes impractical as it is not always feasible to have human experts predetermine these parameters. In response, approaches such as SubgraphX~\cite{yuan2021explainability}, GStarX~\cite{zhang2022gstarx}, and DnX~\cite{pereira2023distill} control the ratio of nodes or edges in the graph instances to form subgraph explanations. However, this involves specifying a fixed ratio applied uniformly to all instances in the datasets. Our study reveals that such a one-size-fits-all ratio may not be widely applicable, as instances in the datasets may require explanations with varying ratios of nodes or edges.

\subsection{Subgraph Inducing Technique}\label{sec:subinduce}

We first provide illustrative examples where edge-induced techniques can provide more intuitive and exhaustive subgraph explanations than node-induced techniques. As shown in \cref{fig:node_edge}(a), if nodes that are not directly neighboring each other are selected, determining the important subgraph structure becomes non-trivial. Taking several ``isolated'' nodes as an implicit representation of the explanatory subgraph loses the inherent intuitive benefits of subgraph-level explanations. In contrast, when edges are selected, the corresponding edge endpoints are naturally selected as well. Therefore, the important subgraph structure is naturally identified. 
Also, producing subgraph-level explanations via node selection may fail to identify some candidate subgraph structures. Taking \cref{fig:node_edge}(b) as an example, if three nodes are connected in the original graph, the underlying triangle connecting the three nodes will be selected to induce a subgraph-level explanation, whereas the true explanation might be the angle-shaped subgraph highlighted in the bottom. Such subgraph selection dilemma based on node selection can be naturally tackled via edge selection methods. This observation is aligned with \cite{10.1145/3447548.3467283}, which points out that highlighting only the nodes is insufficient for providing comprehensive explanations. 

Next, we formally define the \emph{intuitiveness} and the \emph{exhaustiveness} in producing the subgraph-level explanations. We then propose theorems to determine the most effective subgraph inducing technique in both aspects, considering options such as inducing by nodes, edges, or a combination of both nodes and edges. Proofs of all the theorems can be found in \cref{app:proof}. 

\begin{definition}[\bf Intuitiveness of Subgraph-Level Explanations]
    \label{def:intuitiveness}
    The intuitiveness $\mathcal{I}(S)$ of a subgraph-level explanation $S$ is defined as follows: $\mathcal{I}(S)=\frac{C_S}{C}$, where $C$ refers to the number of disconnected components in the explanation $S$, $C_S$ refers to the number of disconnected \emph{subgraph} components in $S$. We define that a disconnected component $G'=(V',E')$ is said to be a disconnected subgraph component iff $|V'|>0$ and $|E'|>0$. 
\end{definition}

\begin{definition}[\bf Exhaustiveness of Subgraph-Level Explanation Inducing Techiniques]
    \label{def:exhaustiveness}
    The exhaustiveness $\mathcal{X}({\mathcal{T}|G})$ of a subgraph-level explanation inducing techinque $\mathcal{T}$ on the corresponding data instance $G=(V,E)$ is defined as follows: $\mathcal{X}({\mathcal{T}|G})=\frac{\mathcal{T}(G)}{C_S}$, where $C_S$ refers to the number of disconnected subgraph components enumerated in $G$, and $\mathcal{T}(G)$ refers to the number of disconnected subgraph components that can be induced by $\mathcal{T}$. 
\end{definition}

\begin{definition}[\bf Node-Induced Subgraph-Level Explanations]
    \label{def:nd_sub}
    Let $G=(V,E)$ denote the data instance, $V_S\subseteq V$ be the node subset to induce the subgraph-level explanation $S$. The node-induced subgraph is defined as $S=G[V_S]=(V_S,E'_S)$, where $E'_S\coloneqq\{\{u,v\}\in E:u,v\in V_S\}$.
\end{definition}

\begin{definition}[\bf Edge-Induced Subgraph-Level Explanations]
    \label{def:ed_sub}
    Let $G=(V,E)$ denote the data instance, $E_S\subseteq E$ be the edge subset to induce the subgraph-level explanation $S$. The edge-induced subgraph is defined as $S=G[E_S]=(V'_S,E_S)$, where $V'_S\coloneqq\{ u,v\in V:\{u,v\}\in E_S\}$.
\end{definition}

\begin{definition}[\bf Node-and-Edge-Induced Subgraph-Level Explanations]
    \label{def:nded_sub}
    Let $G=(V,E)$ denote the data instance, $V_S\subseteq V$ be the node subset and $E_S\subseteq E$ be the edge subset to induce the subgraph-level explanation $S$. The node-and-edge-induced subgraph is defined as $S=G[V_S,E_S]=(V'_S,E'_S)$ where $V'_S\coloneqq V_S \cup \{u,v\in V:\{u,v\}\in E_S\}$ and $E'_S\coloneqq E_S \cup \{\{u,v\}\in E:u,v\in V_S\}$.
\end{definition}

\begin{theorem}
    \label{thm:edge_intuitive}
    Given a graph $G=(V,E)$, an edge-induced subgraph-level explanation $G[E_S]$, a node-induced subgraph-level explanation $G[V_S]$, and a node-and-edge-induced subgraph-level explanation $G[V_S, E_S]$. The following inequalities on the intuitiveness of these explanations always hold, for any $V_S\subseteq V$ and $E_S,E'_S\subseteq E$: 
    $$\mathcal{I}(G[E_S])\geq \mathcal{I}(G[V_S]),$$
    $$\mathcal{I}(G[E_S])\geq \mathcal{I}(G[V_S,E'_S]).$$ 
\end{theorem}

\begin{theorem}
    \label{thm:edge_exhaustive}
    Given a graph $G=(V,E)$, a node-based subgraph inducing algorithm $\mathcal{T}_{\text{node}}$, an edge-based subgraph inducing algorithm $\mathcal{T}_{\text{edge}}$, and a node-and-edge-based subgraph inducing algorithm $\mathcal{T}_{\text{node-and-edge}}$. The following inequality and equation on the exhaustiveness of these subgraph inducing techniques always hold: 
    $$\mathcal{X}(\mathcal{T}_{\text{edge}}|G)\geq \mathcal{X}(\mathcal{T}_{\text{node}}|G),$$
    $$\mathcal{X}(\mathcal{T}_{\text{edge}}|G)= \mathcal{X}(\mathcal{T}_{\text{node-and-edge}}|G).$$ 
\end{theorem}

It is worth noting that $\mathcal{X}(\mathcal{T}_{\text{edge}}|G)= \mathcal{X}(\mathcal{T}_{\text{node-and-edge}}|G)$ since we can consider $\mathcal{T}_{\text{edge}}$ as the special case of $\mathcal{T}_{\text{node-and-edge}}$, where the vertex set $V_S$ fed to $\mathcal{T}_{\text{node-and-edge}}$ is set to $V_S=\O$. However, in real-world scenarios, the explanation vertex set is typically not empty, which poses a risk of failure in identifying the bottom subgraph as illustrated in \cref{fig:node_edge}(b) using the node-and-edge-based inducing algorithm. Therefore, by 
\cref{thm:edge_intuitive} and \cref{thm:edge_exhaustive}, the edge-induced subgraph-level GNN explanations are more comprehensive in the perspectives of intuitiveness and exhaustiveness, compared with the node-induced or node-and-edge-induced subgraph-level explanations.  

\subsection{Size of the Subgraph Explanations}
\label{sec:size_subgraph_exp}
As discussed earlier in \cref{sec:declarations}, presumptions about the number of nodes or edges in subgraph explanations may be ineffective when applied to real-world datasets. Many existing approaches attempt to address this issue by pre-specifying the \emph{sparsity} of the subgraph-level explanations.

\begin{definition}[\bf Sparsity]
    \label{def:sparsity}
    Let $S$ denote the subgraph-level explanation for a graph instance $G=(V,E)$. The sparsity of the explanation $S$ is defined as: $Sparsity ({S}|{G})= 1- \frac{|S|}{|G|}, $
    where $|S|$ and $|G|$ refer to the number of nodes \emph{or} edges in $S$ and $G$. 
\end{definition}

However, controlling the sparsity of these explanations still assumes a certain size for the explanations. This one-size-fits-all ratio may not be universally applicable to all instances in the datasets. For example, in a task that identifies whether there is at least a ``-NO2'' group in the molecules, the true explanation sizes of the p-dinitrobenzene and the nitrobenzene, as illustrated in \cref{fig:node_edge}(c,d), are different. As a result, using sparsity to determine the size of explanations may not be effective. Therefore, it is crucial for the GNN explanation techniques to determine the optimal explanation size for each individual graph. 

The faithfulness of the GNN explanations is commonly assessed using the $Fidelity^+$ and $Fidelity^-$ metrics~\cite{haoyuansurvey2022, GNNBook2022}, which may help to determine the optimal explanation size. We formally define the subgraph-level Fidelity as follows.  

\begin{definition} [\bf Subgraph-Level Fidelity]
    \label{def:fidelity}
    Let $S=(V_S,E_S)$ denote the subgraph-level explanation for a graph instance $G=(V,E)$ on the GNN classifier $\phi(\cdot)$, where $V_S\subseteq V, E_S\subseteq E$. The subgraph-level $Fidelity^+$ of the explanation $S$ is defined as: 
    $$Fidelity^+(S|G)=\phi(G)_y-\phi(G[E \!\setminus\! E_S])_y,$$ 
    where $y$ is the original prediction of the GNN $\phi$ on the graph $G$, $G[E\!\setminus\!E_S]$ refers to the subgraph induced by $E\!\setminus\! E_S$ using the edge-based subgraph inducing technique. Similarly, the subgraph-level $Fidelity^-$ is defined as:
    $$Fidelity^-(S|G)=\phi(G)_y-\phi(S)_y.$$
\end{definition}

Intuitively, $Fidelity^+$ studies the prediction change when the explanation subgraph is removed, while $Fidelity^-$ studies the prediction change when only the explanation subgraph is retained. We calculate $Fidelity^+$ by the subgraph induced by the edge set $E\!\setminus\!E_S$, as opposed to removing $V_S$ and $E_S$ from $G$. This choice is made because removing $V_S$ may lead to edges with missing endpoint nodes in the remaining graph, which could be unnatural for GNNs and potentially cause unexpected behaviors. Both $Fidelity^+$ and $Fidelity^-$ represent the prediction probability change. Higher $Fidelity^+$ and lower $Fidelity^-$ performance indicate that more discriminative subgraph-level explanation is identified. We have the following proposition of these metrics and the optimal size of the subgraph-level explanations. Since this proposition is obvious, we omit the proof. 

\begin{proposition}
    \label{prop:overall_fidelity}
    Given a graph $G=(V,E)$ and a GNN classifier $\phi(\cdot)$, there exists an optimal edge sparsity $\widehat{Sparsity}^{E}(S|G)\in[0,1]$ of the subgraph-level explanation $S$ that maximizes $Fidelity^+(S|G)-Fidelity^-(S|G)$. 
\end{proposition}

Determining $\widehat{Sparsity}^{E}(S|G)$ is challenging, as larger explanation subgraphs do not necessarily lead to better fidelity performance. For example, let ``-NO2'' and ``carbon ring'' be the defining structures for classes $a$ and $b$ respectively of a binary classification problem. Consider the graph in \cref{fig:node_edge}(c) and the metric $Fidelity^+$. Removing both ``-NO2'' groups will certainly result in a dramatic drop in the prediction probability for class $a$. 
However, if we further lower the sparsity by removing more edges, some edges in the carbon ring will be removed, and thus the probability of class $b$ will decrease, which may result into an increase in the probability of class $a$. 
This means that larger subgraph-level explanations may lead to less optimal fidelity performance. Therefore, it is vital for the subgraph-level explainers to be able to determine the optimal sparsity. By \cref{prop:overall_fidelity}, we dertermine the optimal sparsity by the performance of $Fidelity^+(S|G)-Fidelity^-(S|G)$ in our paper. 

\section{Linear-Complexity Search over Edge-induced Subgraphs}
\label{sec:method}

\black{Based on observations in \cref{sec:declarations}, a comprehensive subgraph-level GNN explainer should induce the subgraph-level explanations by edges, and determine the optimal sparsity for each data sample individually.}
Ideally, one could ascertain the optimal sparsity for a given data instance by exhaustively enumerating all edge subsets and select the edge-induced subgraph with the highest fidelity performance. However, due to its exponentially growing computational cost, such enumeration is impractical in real-world applications. 
To this end, we propose an {\model} (\smodel) to produce the subgraph-level explanations \black{in linear time complexity. The intuition is to selectively use only the ``important'' edges to induce subgraphs, which prunes the search space. For example, as we discussed in~\cref{sec:size_subgraph_exp}, the edges in ``-NO2'' are important edges for class $a$, while the ones in the carbon ring are unimportant. }

\begin{figure}[t]
\begin{center}
    \vskip 0.1in
    \centerline{\includegraphics[width=\columnwidth]{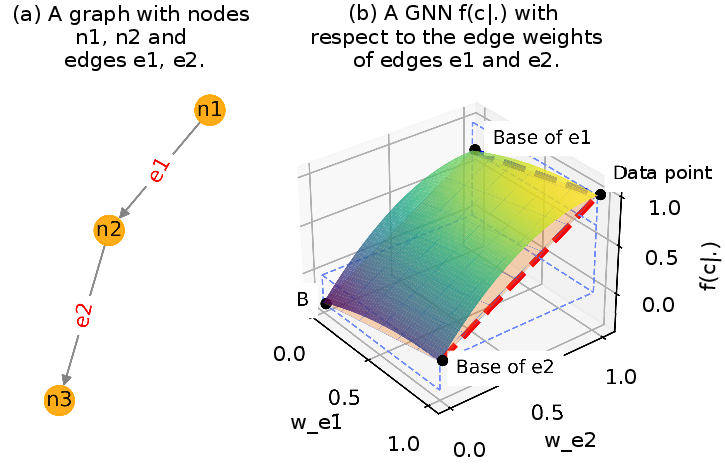}}
    \caption{An example illustrating \emph{edge score approximation}, where $w_{e1}$ and $w_{e2}$ are the edge weights, $c$ is a class of the GNN.}
    \label{fig:intro}
\end{center}
\vskip -0.3in
\end{figure}

{\smodel} achieves the efficiency by using a two-phase scheme to effectively prune the candidate subgraphs that have to be evaluated.
In the first phase, we efficiently approximate each edge's importance by the gradients of the latent lines from the baseline inputs to the original data, with a concept of \emph{Linear Gradient}, which is distinct from the conventional gradient approaches as we will later show in this section. 
This phase results in $O(|E|)$ complexity.
In the second phase, we sort edges by the importance values assigned to them, and induce a subgraph using top-$k$ edges, letting $k$ iterate through $\{1,\dots,|E|\}$. As a result, we obtain $|E|$ candidate subgraph explanations instead of exponentially many. Then, we evaluate all these candidates to find the subgraph explanation that optimizes the subgraph-level fidelity.
Therefore, our design is very \emph{efficient}, with an overall $O(|E|)$ time complexity for the two phases while being \emph{training-free}. The code can be found in the supplementary material. We will explain our approach in detail in the remainder of this section. 


As discussed in \cref{sec:intro}, the explainers that require auxiliary models may introduce non-deterministic behaviors over different runs. Therefore, we avoid training a secondary model in our approach. Instead, we find the importance of edges by constructing latent lines connecting base points to the original data points in space.

As illustrated in \cref{fig:intro}, consider a graph instance $G=(V,E)$ in the dataset, which is classified to Class $c$ by the GNN $\phi(\cdot)$, and we aim to approximate the importance of the edge set $E_t$. We firstly find the ``data point'' in the latent space, which represents the GNN's prediction on the target class $c$ with respect to the weights of all edges in $G$. For each arbitrary edge $e_i\in E$, the edge weight $0\leq w_{e_i}\leq 1$ if $G$ is a weighted graph and $w_{e_i}=1$ if $G$ is unweighted. 
Then, we locate the ``base point'' of $E_t\subseteq E$ in the latent space, which denotes the ``base'' representation of $E_t$. For example, in \cref{fig:intro}(b), if $E_t=\{e_1\}$, then the base point would be the ``Base of e1'' as shown in the figure. As another example, if $E_t=\{e_1,e_2\}$, then the base point of it would be the Point B. The base edge weight could be precisely assigned to $0$ as it denotes a complete absence of signal. This strategy is consistent with~\citet{sundararajan2017axiomatic}. 

After obtaining the representations of the data point and the base point of the target edge set, we construct a line connecting the two points. Next, we use the gradient of this line to approximate the importance of the target edge set. Specifically, the importance of the edge set $E_t$ at the target class $c$ is calculated by 
\begin{equation}
    \label{eq:edge_set_import}
    s(c|E_t) = \frac{\phi(c|\mathbf{A},\mathbf{X}) - \phi(c|\mathbf{A}^{t}, \mathbf{X})} {|\mathbf{A}-\mathbf{A}^t|},
\end{equation}
where $\mathbf{X}$ is the node features, $\mathbf{A}$ is the adjacency matrix, $\mathbf{A}^t$ is the adjacency matrix with the edges in $E_t$ assigned to the corresponding base weights: 
\begin{equation}
    \label{eq:A^t}
    \mathbf{A}^t_{ij}=
        \begin{cases}
        w_{ij,base} & \text{ if } \{v_i,v_j\} \in E_t,\\ 
        \mathbf{A}_{ij} & \text{ if }  \{v_i,v_j\} \notin E_t \text{ and } \{v_i,v_j\}\in E,\\ 
        0 & \text{ if } \{v_i,v_j\}\notin E. 
        \end{cases}
\end{equation}
The denominator of Equation~\eqref{eq:edge_set_import} $|\mathbf{A}-\mathbf{A}^t|$ refers to the distance between the data point and the base point in the latent space. Using Equation~\eqref{eq:edge_set_import}, we can determine the importance of each edge $e_i\in E$ by setting $E_t=\{e_i\}$. In the undirected graphs, each edge $e_i$ is represented by two opposite-direction edges $e_{i,fwd}$ and $e_{i,rvs}$ hence we can obtain the importance of $e_i$ by letting $E_t=\{e_{i,fwd},e_{i,rvs}\}$. 

Our Linear Gradients approach is distinct from the conventional gradient-based methods, like Grad-CAM~\cite{selvaraju2017grad}, DeepLift~\cite{shrikumar2017learning} and Integrated Gradients (IG)~\cite{sundararajan2017axiomatic}. The conventional approaches rely on the gradients that measure the
local sensitivity at the test point, which are susceptible to the saturation problem, leading to vanishing gradients and hence vulnerable to input noise. On the contrary, our approach utilizes the base point to obtain the global importance of an edge rather than a local sensitivity. In particular, Grad-CAM and DeepLift face challenges when applied to edges. As discussed in \cref{sec:declarations}, edge-induced subgraph explanations are more comprehensive, making Grad-CAM and DeepLift less preferred for inducing subgraph-level explanations. IG and our approach share several similarities, including the strategy of selecting base points. However, IG is sensitive to the integral paths. Additionally, due to the high cost of obtaining gradients at all points along the path, IG approximates the integral by summing gradients at a few points, introducing potential errors. Please refer to \cref{app:related_gradient} for detailed discussion. We compare our Linear Gradients approach with several gradient-based approaches to demonstrate the efficacy of our design in \cref{sec:experim}. 

\begin{algorithm}[t]
   \caption{Linear-Complexity Search for Subgraph}
   \label{alg:linear_search}
    \begin{algorithmic}
       \STATE {\bfseries Input:} GNN $\phi(\cdot)$, original graph $G=(V,E)$, ranked edges $\hat{E}$ (decending by importance).
       \STATE Initialize $bestScore=-inf$, $S= None$
       \FOR{$i=2$ {\bfseries to} $|E|-1$}
       \STATE $S = G[\hat{E}[:i]]$, $s(c, G|S) = Fidelity^+-Fidelity^-$
       \IF{$bestScore < s(c, G|S)$}
       \STATE $bestScore = s(c, G|S) $,  $\hat{S} = S$
       \ENDIF
       \ENDFOR
       \STATE {\bfseries Output:} Subgraph-level explanation $\hat{S}$
    \end{algorithmic}
\end{algorithm} 

\begin{figure*}[t]
\begin{center}
\vskip 0.05in
\centerline{\includegraphics[width=0.85\textwidth]{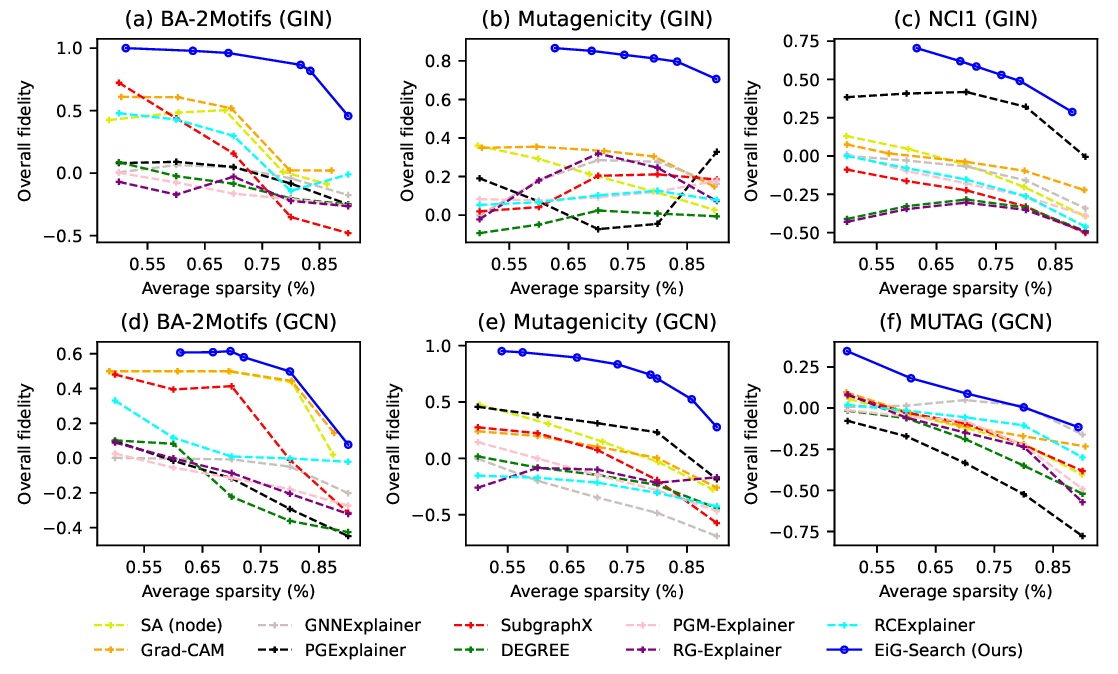}}
\vskip -0.15in
\caption{Overall Fidelity at different levels of average sparsity using {\smodel} and a number of baselines. Higher is better.} 
\label{fig:gc_fidelity}
\end{center}
\vskip -0.3in
\end{figure*}

\begin{table*}[t]
\caption{Efficiency study over different methods on the Mutagenicity dataset.}
\vskip 0.05in
\label{tab:efficiency}
\begin{center}
\scalebox{0.76}{
\begin{tabular}{l|cccccccc}
\toprule
\multirow{2}{*}{Method} & PG & RG-  & RC & GNN  & \multirow{2}{*}{SubgraphX} & \multirow{2}{*}{DEGREE}   & PGM-  & \multirow{2}{*}{\smodel} \\ &Explainer&Explainer& Explainer
 & Explainer&&& Explainer&\\
\midrule
Train Time     & 977.0$\pm$127.5s  & 6359.9$\pm$1257.0s    & 76229.0$\pm$3569.7s & -         & -         & -         & -            & - \\
Explain Time   & 0.03$\pm$0.01s   & 0.03$\pm$0.01s  & 0.07$\pm$0.03s & 1.44$\pm$ 0.09s      & 419.8$\pm$655.5s     & 0.53$\pm$0.34s     & 0.86$\pm$0.76s        & 0.14$\pm$0.01s \\
\bottomrule
\end{tabular}
}
\end{center}

\vskip -0.1in
\caption{Efficiency study over different efficient methods.}
\label{tab:full_efficiency}
\begin{center}
\scalebox{0.8}{
\begin{tabular}{l|cccccc}
\toprule
& BA- & BA- & Tree- & BA- & \multirow{2}{*}{MUTAG} & \multirow{2}{*}{Mutagenicity}\\
& Shapes & Community & Grid & 2Motifs & &\\
\midrule
GNNExplainer~\cite{ying2019gnnexplainer} & 0.65$\pm$0.05s&0.78$\pm$0.05s&0.72$\pm$0.06s&1.16$\pm$0.10s&0.43$\pm$0.03s&1.44$\pm$0.09s\\
PGExplainer~\cite{luo2020parameterized} &0.181$\pm$0.04s& 0.0495$\pm$0.01s & 0.215$\pm$0.07s & 0.306s$\pm$0.07&0.138$\pm$0.03s&0.274$\pm$0.08s\\
DEGREE~\cite{feng2022degree} &0.44$\pm$0.20s &1.02$\pm$0.35s & 0.37$\pm$0.06s & 0.575$\pm$0.11s &0.83$\pm$0.45s & 0.53$\pm$0.34s \\
{\smodel} (Ours) & \textbf{0.006$\pm$0.000s}& \textbf{0.007$\pm$0.000s}& \textbf{0.003$\pm$0.000s}& \textbf{0.089$\pm$0.01s}&\textbf{0.07$\pm$0.01s}& \textbf{0.14$\pm$0.01s}\\
\bottomrule
\end{tabular}
}
\end{center}
\vskip -0.15in
\end{table*}

Once we obtain the edge importance of all the edges in a graph by Equation~\eqref{eq:edge_set_import}, we can rank the edges through the importance scores. As discussed in \cref{sec:size_subgraph_exp}, simply having the rank of important features is insufficient, we also need to determine the optimal sparsity of an explanation. Rather than employing an expensive enumeration of edge-induced subgraphs, our approach utilizes a more efficient linear-time search over subsets of the ranked edges. This search is guided by the fidelity performance of the edge-induced subgraph-level explanations. The pseudo code of our Linear-Complexity Search is presented in \cref{alg:linear_search}. Thus, we can obtain the optimal subgraph-level explanation $\hat{S}$ of a graph $G$ with an optimal sparsity by picking the $S$ that gives the best overall score $s(c, G|S)$. 
\begin{figure*}[t]
\begin{center}
\centerline{\includegraphics[width=0.9\textwidth]{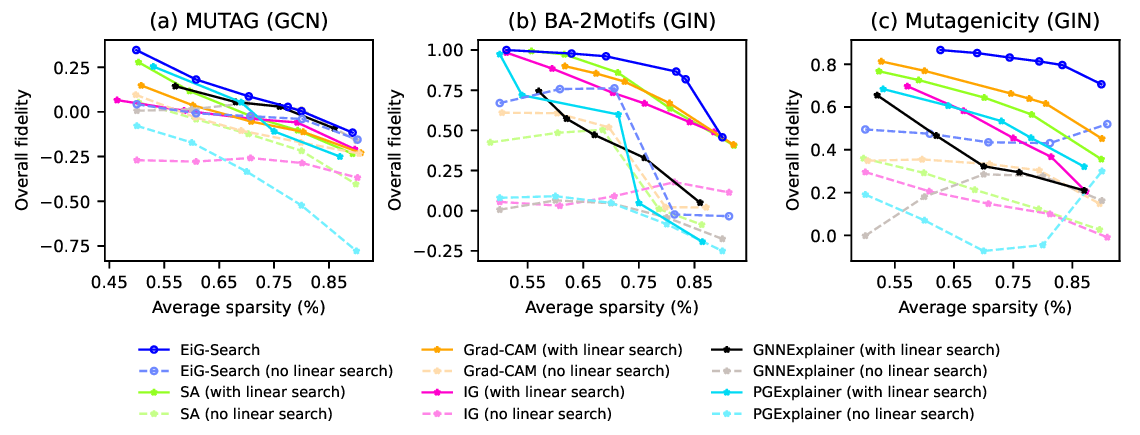}}
\vskip -0.1in
\caption{Comparsion between the baselines and {\smodel} after applying \emph{Linear-Complexity search}.}
\label{fig:ls} 
\end{center}
\vskip -0.3in
\end{figure*}

\section{Experiments}
\label{sec:experim}

In this section, we perform empirical evaluations of our proposed {\smodel}. We mainly consider the following three sets of experiments. \emph{Firstly}, to validate the overall effectiveness of our two-phase pipeline design, we compare the faithfulness of subgraph-level explanations generated by existing explainers with those produced by {\smodel}. \emph{Secondly}, we highlight the effectiveness of each phase by integrating our Linear-Complexity Search algorithm with existing explainers that generate node-level or edge-level explanations. On one hand, this augmentation allows us to assess whether our linear-time search can enhance the performance of existing methods. On the other hand, by comparing the augmented baselines to {\smodel}, we investigate whether our Linear Gradients method provides a better approximation of edge importance compared to existing approaches. \emph{Thirdly}, we perform empirical time analysis to showcase the efficiency of {\smodel}. 

\paragraph{Faithfulness of the entire framework.} We use the \emph{subgraph-level fidelity} metric mentioned in ~\cref{prop:overall_fidelity} with \cref{def:fidelity}, i.e., the overall fidelity calculated by subtracting $Fidelity^-$ from $Fidelity^+$, to evaluate the faithfulness of the subgraph-level GNN explanations. We conduct experiments both on the synthetic dataset BA-2Motifs~\cite{luo2020parameterized}, and the real-world datasets MUTAG~\cite{debnath1991structure}, Mutagenicity~\cite{kazius2005derivation}, NCI1~\cite{wale2006comparison}. While our approach can generalize to any type of GNN, we consider two popular variants: Graph Convolutional Networks (GCN)~\cite{kipf2017semisupervised} on BA-2Motifs, Mutagenicity and MUTAG as well as Graph Isomorphism Networks (GIN)~\cite{xu2018powerful} on BA-2Motifs, Mutagenicity, and NCI1. We conduct extensive experiments to compare our method with the state-of-the-art methods including the gradient-based SA~\cite{baldassarre2019explainability} and Grad-CAM~\cite{pope2019explainability}, perturbation-based GNNExplainer~\cite{ying2019gnnexplainer} and PGExplainer~\cite{luo2020parameterized}, search-based SubgraphX~\cite{yuan2021explainability}, decomposition-based DEGREE~\cite{feng2022degree}, surrogate-based PGM-Explainer~\cite{vu2020pgm}, RL-based RG-Explainer~\cite{shan2021reinforcement} and decision boundary-based RCExplainer~\cite{bajaj2021robust}. We run experiments using the open-source implementations of these works. All the baselines necessitate the pre-specification of subgraph-level explanation sizes. To facilitate a fair comparison with these baselines, we adopt the setup outlined in~\citet{bajaj2021robust}, where the fidelity results are evaluated and compared across a range of edge sparsity levels. 
Among the baselines, SubgraphX, DEGREE, RG-Explainer were designed to provide subgraph-level explanations, while other baseline methods only provide node-level or edge-level explanations. We induce subgraph-level explanations with the explanations produced by these node-level and edge-level explainers according to \cref{def:nd_sub} and \ref{def:ed_sub}. In particular, SA, Grad-CAM, PGM-Explainer generate node-induced subgraph explanations, while GNNExplainer, PGExplainer, RCExplainer generate edge-induced subgraph explanations in our experiments. 
The details of model configurations and datasets are provided in \cref{app:stats}.

The overall fidelity results are presented in \cref{fig:gc_fidelity}. The $Fidelity^+$ and $Fidelity^-$ results can be found in \cref{app:fidelity+-}. Our proposed {\smodel}, along with SA, Grad-CAM, SubgraphX and DEGREE, belongs to the category of training-free methods. This distinguishes them from GNNExplainer, PGExplainer, PGM-Explainer, RG-Explainer and RCExplainer, which require training. As shown in \cref{fig:gc_fidelity}, across various datasets, training-free methods exhibit greater consistency in performance compared to the training-required PGExplainer. While PGExplainer achieves notably strong results on NCI1 (GIN), its performance is less effective on MUTAG (GCN), where it demonstrates the poorest fidelity results. Our proposed {\smodel} outperforms both node-induced and edge-induced baselines in fidelity across all datasets, showcasing the strength of the entire design. The \emph{qualitative} results can be found in \cref{app:quality}.

\paragraph{Effectiveness of each phase.} We further study the empirical performance of the Linear Gradients and Linear-Complexity Search respectively, which together constitute our proposed {\smodel}. We integrate the Linear-Complexity Search algorithm with both typical training-requiring node-level or edge-level baselines, such as GNNExplainer and PGExplainer, and various training-free baselines, including SA, Grad-CAM, and Integrated Gradients (IG). We generate node-induced subgraph explanations with SA and Grad-CAM, and edge-induced subgraph explanations with GNNExplainer, PGExplainer and IG. We evaluate the overall fidelity on MUTAG, BA-2Motifs, Mutagenicity. 

The results are presented in \cref{fig:ls}. For a more intuitive comparison between {\smodel} and each baseline, please refer to \cref{app:abs_detail}. The performance of both node-level and edge-level baselines is consistently enhanced when augmented with our Linear-Complexity Search. This observation strongly underscores the effectiveness of Linear-Complexity Search in improving subgraph-level explanations for GNNs by finding the diverse explanation sizes of various data instances. Furthermore, it can be seen that {\smodel} consistently outperforms the augmented node-level and edge-level baselines, providing evidence that our Linear Gradients method offers a more accurate approximation of edge importance. It is also noteworthy that despite Integrated Gradients (IG) aiming to approximate global edge importance similar to our approach, it consistently performs worse than our Linear Gradients. It even fares slightly worse than conventional gradient approaches like SA and Grad-CAM. This discrepancy may be attributed to IG's sensitivity to the integral paths. If the integral paths are not faithfully chosen, optimal results are not guaranteed. Determining accurate integral paths, especially in high-dimensional spaces, poses a significant challenge. Additionally, IG can only approximate the importance of individual edges and not edge sets, as our Linear Gradients can. 
In undirected graphs, where each undirected edge is usually represented by two opposite-direction edges between the endpoints, failure to treat these opposite-direction edges as an entire component may be a contributing factor impacting the performance of IG. 
In summary, our comprehensive experiments demonstrate the effectiveness of both Linear Gradients and Linear-Complexity Search in {\smodel}.  

\paragraph{Efficiency.} \cref{tab:efficiency} shows the average computation time for providing the post-hoc explanation on the graph sample from the Mutagenicity dataset with GCN. Our approach has the fastest explaining time compared to the presented recent baselines that do not require explainer training. Note that PGExplainer, RG-Explainer, and RCExplainer all require additional training time for the explainer, which is costly. 
Next, we present a detailed efficiency comparison between {\smodel} and the baseline GNN explanation approaches, including the most efficient training-required approach PGExplainer, and two efficient approaches GNNExplainer and DEGREE. To calculate the average time required for producing an explanation, we consider the sum of the training time and the inference time on all the data samples for the baselines that require training. This sum is then divided by the number of data samples. 
For example, let's consider PGExplainer on the BA-2Motifs dataset with 1000 data samples. It takes 225.5 seconds to train the explainer, and 0.08 seconds to infer the explanation for a single data sample. In this case, the average time to provide an explanation on this dataset is $\frac{225.5s}{1000}+0.08s=0.306s$. As shown in \cref{tab:full_efficiency}, {\smodel} consistently outperforms the baseline approaches in terms of efficiency across all datasets, including both node classification and graph classification tasks. This emphasizes the superior efficiency of our approach in generating subgraph-level explanations.

\paragraph{Other tasks.} We also perform node-level experiments on three popular datasets with the AUC metric. Due to space limit, the results are shown in Appendix~\ref{app:node_class}.

\section{Conclusion}
In this paper, we systematically study the process of generating subgraph explanations for GNNs from the perspectives of subgraph inducing techniques and optimal explanation size. We show the advantage of edge-induced subgraphs and design a simple yet efficient, model-agnostic method to find the optimal subgraph explanation in linear time given a graph instance. 
We empirically demonstrate the effectiveness and efficiency of {\smodel} through extensive experiments.

\section*{Impact Statement}

Our technique aims to contribute to the community's understanding of the decision-making process in GNNs and enhance the reliability of these models. We hope that our approach will be valuable in advancing the field and fostering greater trust and transparency in GNNs. 
Unlike many existing subgraph-level explainers that focus on node-induced subgraph explanations, we demonstrate that edge-induced subgraph explanations offer a more intuitive and exhaustive understanding. Additionally, we emphasize the importance of identifying the explanation size rather than employing a uniform sparsity level for all instances in a dataset. These findings may inspire further exploration within the community on the potential of edge-induced subgraph explanations.
The core concept of our proposed GNN explaining approach, {\smodel}, involves a two-phase process consisting of an edge-ranking algorithm and a linear-time search algorithm. While {\smodel} is known for its efficiency and training-free nature, there remains room for improvement in explanation performance, albeit with a tradeoff in efficiency. In the future, the development of a more accurate edge-ranking algorithm and an advanced subgraph search, guided by edge importance, could be promising directions for further research.

\nocite{ye2023same}

\bibliography{draft}
\bibliographystyle{icml2024}

\newpage
\appendix
\onecolumn

\section{Proof of Theorems}\label{app:proof}
\subsection{Proof of \cref{thm:edge_intuitive}}
\label{app:thm:edge_intuitive}

\begin{proof}
    From the definition, it is obvious that the vertex subset of the edge-induced subgraph-level explanation are exactly all endpoints of the edge subset. In other word, all the disconnected components in the edge-induced subgraph-level explanation are disconnected \emph{subgraph} components. Therefore, the intuitiveness of the edge-induced subgraph-level explanation $G[E_S]$ can be calculated as $\mathcal{I}(G[E_S])=\frac{C_S}{C}=1$. In contrast, in a node-induced subgraph-level explanation, although the vertex subset include all endpoints of the edge subset, it may also include additional vertices. As a result, the disconnected components in these explanations may not be the intuitive disconnected subgraph components. The intuitiveness of the node-induced subgraph-level explanation $G[V_S]$ can be calculated as $\mathcal{I}(G[V_S])=\frac{C_S}{C}\leq1$. When there are additional vertices excluding the endpoints of the edges in $G[V_S]$, we have $\mathcal{I}(G[V_S])<1$. Lastly, in a node-and-edge-induced subgraph-level explanation $G[V_S,E_S]$, all the endpoints of edges in $E_S$ are selected, but there can be additional vertices in $V_S$ that contruct the disconnected node components in $G[V_S,E_S]$. Therefore, similar to the node-induced subgraphs, we have $\mathcal{I}(G[V_S,E_S])\leq1$. Hence, we have proved that for any $V_S\subseteq V$ and $E_S,E'_S\subseteq E$:
    $$\mathcal{I}(G[E_S])\geq \mathcal{I}(G[V_S]),$$
    $$\mathcal{I}(G[E_S])\geq \mathcal{I}(G[V_S,E'_S]).$$ 
\end{proof}

\subsection{Proof of \cref{thm:edge_exhaustive}}
\label{app:thm:edge_exhaustive}

\begin{proof}
    By definition, the exhaustiveness of the subgraph inducing technique applied to the edge-induced subgraph-level explanations is $\mathcal{X}(\mathcal{T}_{\text{edge}}|G)=1$. For the node-based inducing technique, we have $\mathcal{X}(\mathcal{T}_{\text{node}}|G)\leq1$. In particular, we have $\mathcal{X}(\mathcal{T}_{\text{node}}|G)<1$ when there are cycles in $G$. As $T_{\text{node}}(G)$ misses the disconnected subgraph components by removing any one of the edges in a cycle. 
    The node-and-edge-based inducing technique is equivalent to the edge-based inducing algorithm when $|V_S|=\O$. Hence it has the same exhaustiveness as the edge-based subgraph inducing algorithm. 
    Both the node-based technique and the node-and-edge-based technique is able to produce isolated nodes. But it is important to note that isolated nodes are not disconnected subgraph components according to \cref{def:intuitiveness}, hence they will not count into $\mathcal{T}(G)$ or $C_S$. Hence we have proved that 
    $$\mathcal{X}(\mathcal{T}_{\text{edge}}|G)\geq \mathcal{X}(\mathcal{T}_{\text{node}}|G),$$
    $$\mathcal{X}(\mathcal{T}_{\text{edge}}|G)= \mathcal{X}(\mathcal{T}_{\text{node-and-edge}}|G).$$ 
\end{proof}

\section{Further Discussions with Prior Works}\label{app:related_work}
We have briefly reviewed and distinguished the prior works in \cref{sec:intro}, \ref{sec:declarations} and \ref{sec:experim}. In this section, we provide a comprehensive review of related works in the domain of instance-level post-hoc GNN explainability. Recall that our approach is motivated by the aim to introduce a GNN explaining method that is both \emph{training-free} and \emph{efficient}, delivering \emph{intuitive subgraph-level explanations} for graph instances. To elucidate this motivation, we categorize existing approaches based on two criteria: whether the explainer requires training and the technique used for subgraph induction. 

\subsection{Related Works Grouped by Training Requirement} \label{app:related_gradient}

Based on the training requirements, existing instance-level GNN explaining approaches can be divided into two groups: training-free and training-requiring approaches. Training-requiring methods usually train a secondary black-box, which reduces the interpretability and transparency of the explainers. Moreover, the explanations generated by these explainers can be inconsistent across different runs. As pointed out by~\citet{zhao2023towards}, these methods may introduce non-deterministic behaviors even for the same input graph since they require training an auxiliary or secondary model. A lack of consistency will compromise the faithfulness of the explanation as well. 
Approaches including GNNExplainer~\cite{ying2019gnnexplainer}, PGExplainer~\cite{luo2020parameterized}, PGM-Explainer~\cite{vu2020pgm}, GraphMask~\cite{schlichtkrull2020interpreting}, RelEx~\cite{zhang2021relex}, RCExplainer~\cite{bajaj2021robust}, RG-Explainer~\cite{shan2021reinforcement}, Gem~\cite{lin2021generative}, GraphLime~\cite{huang2022graphlime}, GFlowExplainer~\cite{li2023dag}, DnX~\cite{pereira2023distill}, K-FactExplainer~\cite{huang2023factorized}, fall into the training-requiring category. We provide more discussions on these explainers in \cref{app:related_inducing}. In contrast, training-free methods are usually more transparent, making them more reliable. Therefore, it is crucial to design training-free explainers. 

The training-free methods can be further divided into gradient-based methods and search-based methods. 
The gradient-based methods include SA, Guide-BP, LRP~\cite{baldassarre2019explainability}, Grad-CAM, Excitation-BP~\cite{pope2019explainability}, GNN-LRP~\cite{schnake2021higher}, DeepLIFT~\cite{shrikumar2017learning, haoyuansurvey2022}, Interated Gradients (IG)~\cite{sundararajan2017axiomatic}. These methods are originally designed to explain general neural networks like CNNs and adapted for use with GNNs. They are white-box approaches but are noted to suffer from the gradient saturation problem~\cite{shrikumar2017learning, sundararajan2017axiomatic}. Furthermore, these methods have various drawbacks. Grad-CAM obtains the activation map by multiplying two terms. One is the hidden embedding ahead of the classifier layers, the other is the output gradient with respect to the hidden embedding. However, in GNNs, there is only hidden node embedding instead of hidden edge embedding. Thus, CAM-based methods are not applicable to the adjacency matrix, hence are not able to provide intuitive and exhaustive edge-induced subgraph-level explanations as we discussed in \cref{sec:subinduce}. LRP, Excitation-BP and DeepLIFT are decomposition methods. Since layer-wise back-propagation is performed, they require expert knowledge of the original GNNs and need specific designs for various GNN structures. Besides, different from a traditional input of neural networks that appears only once at the first layer of the whole network, the adjacency matrix $\mathbf{A}$ appears in every GNN layer. This makes decomposing the prediction probability to the components in $\mathbf{A}$ a much more complicated problem than decomposing to node-level. GOAt~\cite{lu2024goat} requires expert knowledge to design different attribution equations for different GNN architectures. And our method, {\smodel}, is extremely trivial to implement and can be applied to all the GNNs. IG computes the edge importance by the integral of edge gradients starting from a global base point. However, the performance is sensitive to the integral paths. As it is costly to obtain the gradients at all points on the path, the integral is approximated by the summation of a few gradients along the path, which could bring more error. And our proposed {\smodel} eliminates the need for guessing integral paths. 
The search-based methods including SubgraphX~\cite{yuan2021explainability}, GStarX~\cite{zhang2022gstarx}, SAME~\cite{ye2023same}, are typically utilized to search for the best subgraph-level explanations. Due to the computational challenges associated with searching among the exponential number of candidate subgraphs, these methods often use Monte Carlo Tree Search (MCTS) to speed up the process. However, MCTS introduces randomness, leading to non-deterministic behaviors to the explanations even for the same input graph. Further discussions on these methods are provided in \cref{app:related_inducing}. In contrast, the two-phase design of {\smodel} allows it to be highly efficient while maintaining consistency across various runs. 

\subsection{Related Works Grouped by Subgraph Inducing Technique} \label{app:related_inducing}

The existing approaches can be grouped into node-level, edge-level and subgraph-level methods. SubgraphX~\cite{yuan2021explainability}, RG-Explainer~\cite{shan2021reinforcement}, DEGREE~\cite{feng2022degree}, GStarX~\cite{zhang2022gstarx}, SAME~\cite{ye2023same}, GFlowExplainer~\cite{li2023dag}, DnX~\cite{pereira2023distill} belong to the subgraph-level methods. We surprisingly find that all of these methods choose to induce the subgraph-level explanations by vertices. However, as we discussed in \cref{sec:subinduce}, edge-induced technique, as our method {\smodel} uses, offers more intuitive and exhaustive subgraph-explanations. Additionally, we argue that a single connected subgraph may not always be sufficient to explain the GNN prediction on a graph instance. Take, for instance, a binary graph classification task where the presence of ``-NO2'' designates Class 1; otherwise, it is Class 0. In this scenario, the most accurate subgraph-level explanation for p-dinitrobenzene, as shown in \cref{fig:node_edge}(c), should be the two ``-NO2'' components rather than a single connected component. From this perspective, our proposed {\smodel}, which first identifies the critical edge set and then induces subgraph explanations, is better than the methods that select an anchor point and then grow by neighbors to find a single important connected component, as seen in approaches like RG-Explainer. 

Although node-level and edge-level explanations are less intuitive than subgraph-level explanations, we can induce subgraph explanations with the node or edge importance produced by the node or edge-level explainers. Node-level explainers including Grad-CAM, Excitation-BP~\cite{pope2019explainability}, LRP~\cite{baldassarre2019explainability}, DeepLIFT~\cite{shrikumar2017learning}, PGM-Explainer~\cite{vu2020pgm}, GraphLime~\cite{huang2022graphlime}, can be used to induce high quality subgraph-level explanations in most cases. However, as we have explained in \cref{sec:subinduce}, node-induced subgraph explanations can be less intuitive than the edge-induced subgraph explanations, and the node-level subgraph inducing technique is less exhaustive than the edge-level technique that our method uses. To this end, there are many works that are able to produce edge-level explanations, including GNNExplainer~\cite{ying2019gnnexplainer}, PGExplainer~\cite{luo2020parameterized}, GraphMask~\cite{schlichtkrull2020interpreting}, RelEx~\cite{zhang2021relex}, Gem~\cite{lin2021generative}, RCExplainer~\cite{bajaj2021robust}, SA~\cite{baldassarre2019explainability}, Integrated Gradients (IG)~\cite{sundararajan2017axiomatic}, GOAt~\cite{lu2024goat}. The limitations of these works are discussed in \cref{app:related_gradient}. 

\section{Experimental Details}\label{app:exp_detail} 

\subsection{\emph{Fidelity$^+$} and \emph{Fidelity$^-$} performance on the graph classification tasks.}
\label{app:fidelity+-}

\begin{figure*}[t]
\begin{center}
\centerline{\includegraphics[width=0.9\textwidth]{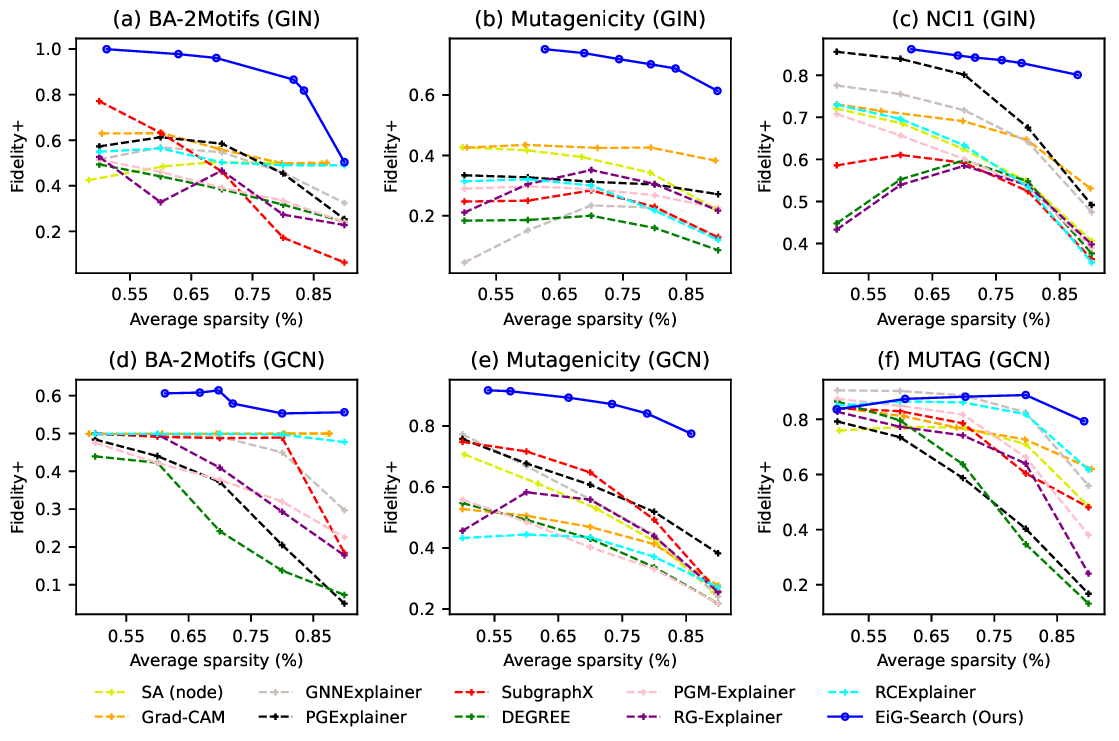}}
\caption{Fidelity+ at different levels of average sparsity. The higher the better. }
\label{fig:gc_fidelity+}
\vskip 0.3in
\centerline{\includegraphics[width=0.9\textwidth]{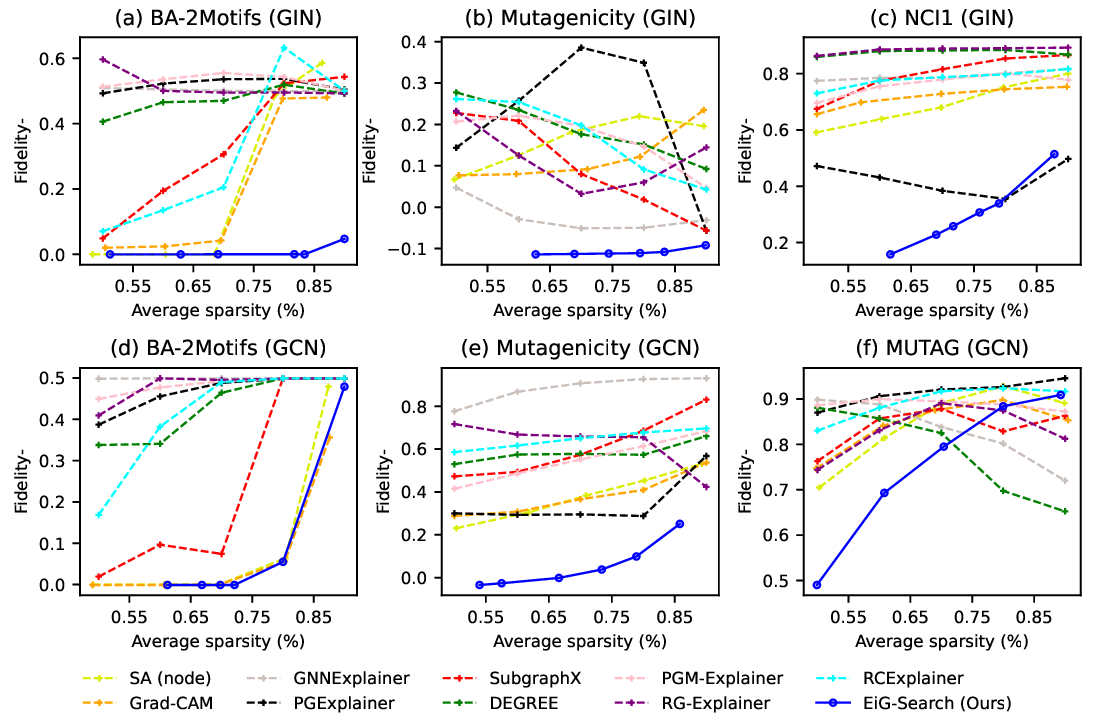}}
\caption{Fidelity- at different levels of average sparsity. The lower the better. }
\label{fig:gc_fidelity-}
\end{center}
\end{figure*}

We report separate $Fidelity^+$ and $Fidelity^-$ results in Figure~\ref{fig:gc_fidelity+} and Figure~\ref{fig:gc_fidelity-}. The results are obtained while optimizing the Overall Fidelity, hence in some datasets, e.g. MUTAG, $Fidelity^+$ does not necessarily decrease as sparsity increases. In this case, the performance that $Fidelity^-$ gains outweighs $Fidelity^+$, resulting in higher Overall Fidelity in Figure~\ref{fig:gc_fidelity}. 

\subsection{Qualitative Results.} 
\label{app:quality}
As demonstrated in Table \ref{tab:quality_detail_ba2motifs} and Table~\ref{tab:quality_detail_other}, {\smodel} excels at identifying significant subgraph structures, such as the "house" and "pentagon" motifs in the BA-2Motifs dataset, the "C-Cl-O" chemical group in the Mutagenicity dataset, and the carbon ring in the NCI1 dataset. On the other hand, the baseline methods like GNNExplainer, PGExplainer, PGM-Explainer, RCExplainer, RG-Explainer, DEGREE, and Integrated Gradients struggle to generate human-interpretable subgraph-level explanations. Moreover, these baseline methods fail to recognize structural and computational equivalents among edges. For instance, RCExplainer marks only one of the "C-O" bonds in Table~\ref{tab:quality_detail_other} as critical, while leaving the other one unmarked, despite both bonds having identical neighborhood information. In contrast, {\smodel} assigns equal importance to edges with identical neighborhood information, selecting all of them as critical when one is selected. This highlights the efficacy of {\smodel} in providing comprehensive subgraph-level explanations.

\subsection{Comparison between {\smodel} and each Augmented Baseline} \label{app:abs_detail}

\begin{figure*}[htp]
\begin{center}
\centerline{\includegraphics[width=0.9\textwidth]{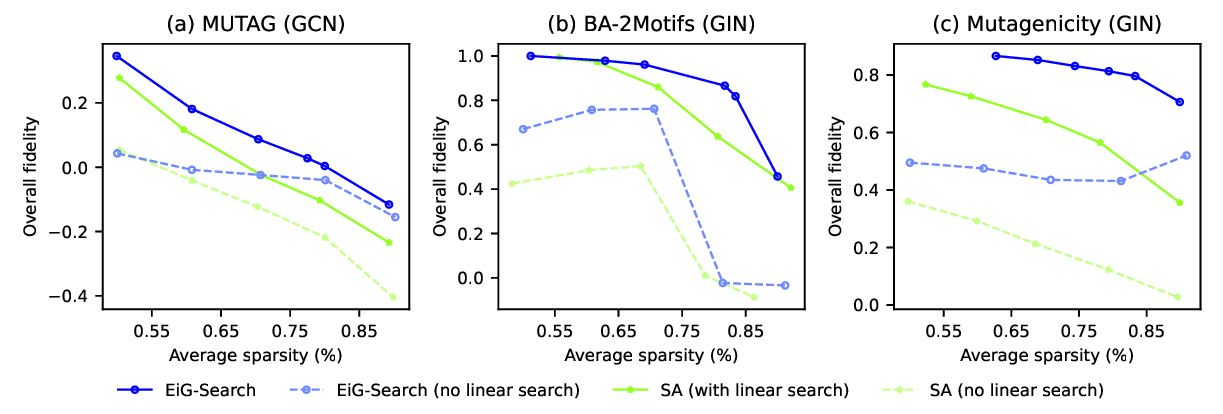}}
\vskip -0.2in
\caption{Comparsion between SA and {\smodel} after applying \emph{Linear-Complexity search}. }
\label{fig:app_sa}
\vskip 0.2in
\centerline{\includegraphics[width=0.9\textwidth]{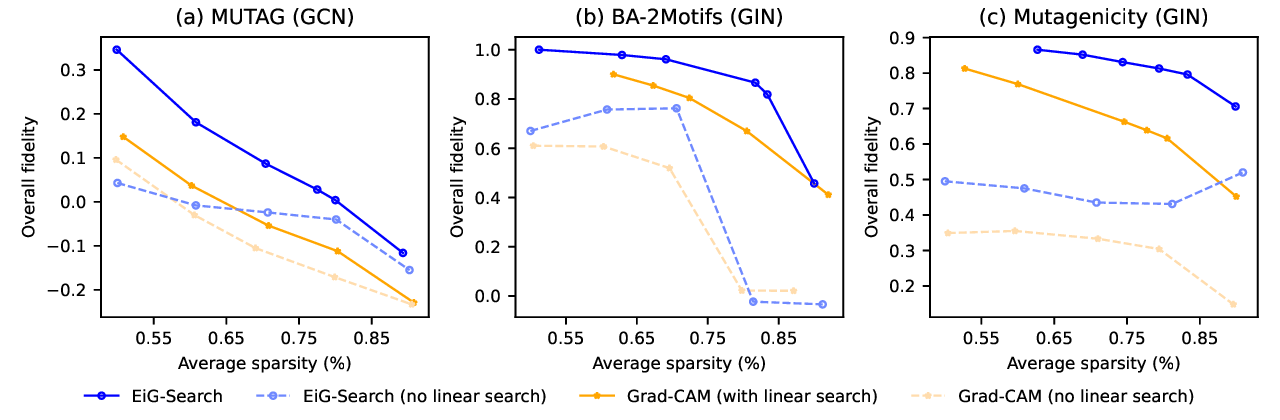}}
\vskip -0.2in
\caption{Comparsion between Grad-CAM and {\smodel} after applying \emph{Linear-Complexity search}}
\label{fig:app_gradcam}
\end{center}
\end{figure*}

\begin{figure*}[htp]
\begin{center}
\centerline{\includegraphics[width=0.9\textwidth]{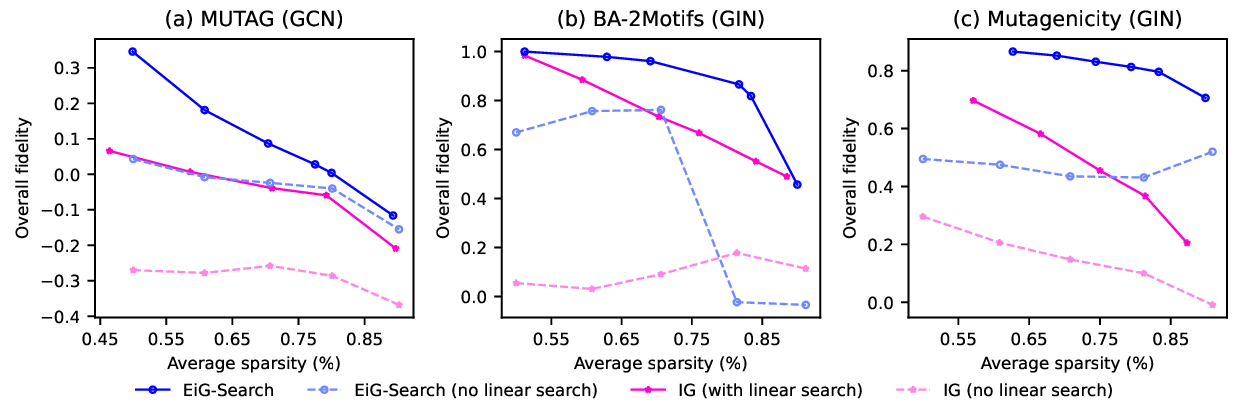}}
\vskip -0.2in
\caption{Comparsion between IG and {\smodel} after applying \emph{Linear-Complexity search}. }
\label{fig:app_ig}
\vskip 0.2in
\centerline{\includegraphics[width=0.92\textwidth]{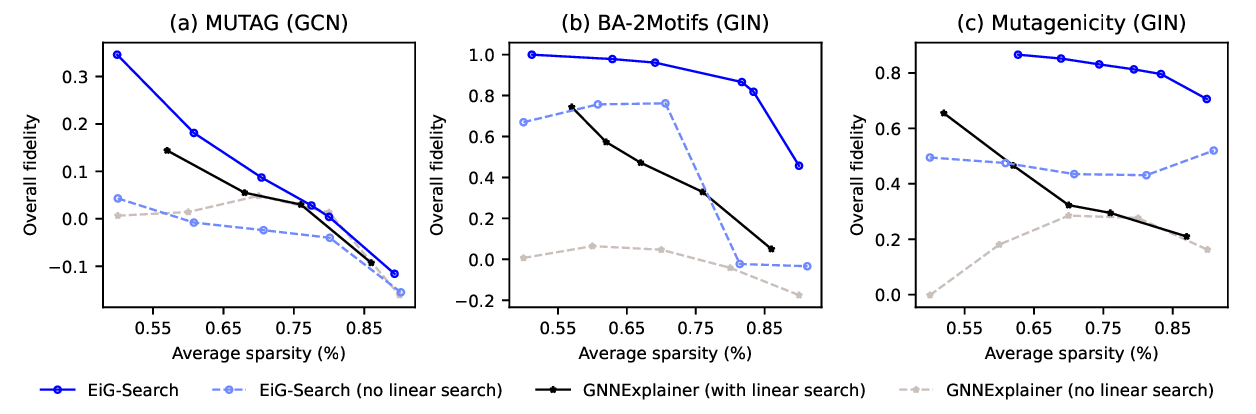}}
\vskip -0.2in
\caption{Comparsion between GNNExplainer and {\smodel} after applying \emph{Linear-Complexity search}}
\label{fig:app_gnn}
\vskip 0.2in
\centerline{\includegraphics[width=0.92\textwidth]{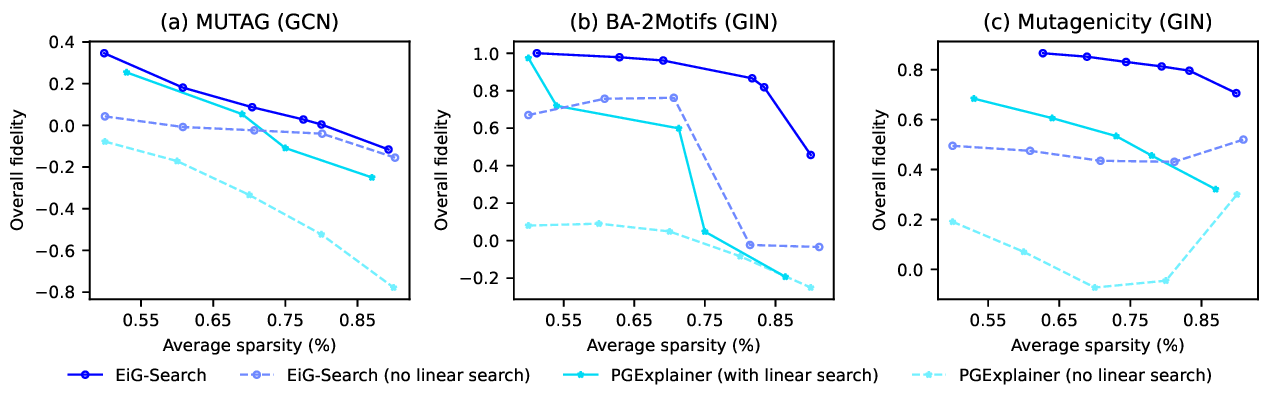}}
\vskip -0.2in
\caption{Comparsion between PGExplainer and {\smodel} after applying \emph{Linear-Complexity search}}
\label{fig:app_pg}
\end{center}
\end{figure*}

\cref{fig:app_sa}-\ref{fig:app_pg} presents a more intuitive comparison between our {\smodel} and each augmented baseline. The baselines are augmented with our proposed search method in \cref{alg:linear_search}. In particular, for IG, we use the straightline path and let the step size be $50$ as suggested in their paper. 

\subsection{More Tasks}\label{app:node_class}
\paragraph{Node-level tasks.} We further conduct experiments on three node-level tasks: BA-Shapes, BA-Community, Tree-Grid~\cite{ying2019gnnexplainer}. For node classification, the node itself is important for its prediction and we cannot remove it from the graph, otherwise, we cannot make a prediction on it or compute fidelity.
However, as highlighted by~\citet{10.1145/3447548.3467283}, when we are able to train an GNN close to its maximum possible performance (e.g. 100\% prediction rate for classification), it is likely that we can use the ground-truth explanations to evaluate the explainers. Therefore, we train GINs to near optimal performance on the node classification datasets, and use the explanation AUC to evaluate the performance of explaining methods, which aligns with \citet{ying2019gnnexplainer}, \citet{luo2020parameterized}, \citet{bajaj2021robust}, \citet{feng2022degree}. 

We compare our method {\smodel}, in particular, the Linear Gradients method, with the state-of-the-art explaining techniques on node classification tasks. We report AUC under the ROC curve in Table~\ref{tab:node_class}. 
The results demonstrate that our approach is very accurate in extracting the optimal explanations on these datasets. Another advantage is that {\smodel} does not require any hyperparameters, and thus is more faithful to the GNNs, unlike PGExplainer, RCExplainer and RG-Explainer, which require tuning hyperparameters for each dataset. 

\begin{table}[h]
\vskip -0.1in
\caption{AUC evaluation on synthetic node classification datasets. The results of baselines are from the original papers. }
\vskip 0.1in
\label{tab:node_class}
\centering
\begin{tabular}{l|ccc}
\toprule
\multirow{2}{*}{Method} & BA- & BA- & Tree-  \\
&Shapes&Community&Grid\\
\midrule
GRAD    & 0.882 & 0.750 & 0.612 \\
ATT    & 0.815 &  0.739   &  0.667    \\
GNNExplainer  & 0.925 & 0.836 & 0.875 \\
PGExplainer  & 0.963 & 0.945 & 0.907 \\
DEGREE  & 0.991 & 0.984    & 0.935     \\
RG-Explainer  & 0.985 & 0.919    & 0.787     \\
RCExplainer & 0.998 & 0.995    &  \textbf{0.995}    \\
\midrule
{\smodel} (ours) & \textbf{0.999} &  \textbf{0.996}  & 0.947     \\
\bottomrule
\end{tabular}
\end{table}

\subsection{Statistics of datasets and GNNs.}
\label{app:stats}

The statistics of datasets and GNNs are presented in Table~\ref{tab:stats}. The GNNs are trained with the following data splits: training set (80\%), validation set (10\%), testing set (10\%). All the experiments are conducted on Intel® Core™ i7-10700 Processor and NVIDIA GeForce RTX 3090 Graphics Card. All the GNNs contain 3 message-passing layers and a 2-layer classifier, the hidden dimension is 32 for BA-2Motifs, BA-Shapes, and 64 for BA-Community, Tree-grid, MUTAG, Mutagenicity and NCI1. 

\begin{table*}[h]
\caption{Statistics of the datasets used and the train/test accuracy of the trained GNNs.}
\label{tab:stats}
\vskip 0.1in
\begin{center}
\begin{sc}
\begin{tabular}{ll|ccc|cccc}
\toprule
 && BA- & BA- & Tree- & BA- & \multirow{2}{*}{MUTAG} & \multirow{2}{*}{Mutagenicity} & \multirow{2}{*}{NCI1} \\
  && Shapes & Community & Grid & 2Motifs & & &\\
\midrule
\multicolumn{2}{l|}{\# Graphs} & 1&1&1& 1,000 &188 &4,337 &4,110 \\
\multicolumn{2}{l|}{\# Nodes (avg)}  & 700 & 1,400 &1,231 &25 & 17.93 & 30.32 & 29.87\\
\multicolumn{2}{l|}{\# Edges (avg)} & 4,110 & 8,920 & 3,410& 25.48& 19.79 & 30.77 & 32.30\\
\multicolumn{2}{l|}{\# Classes } & 4 & 8 & 2& 2&2&2&2\\
\midrule
\multirow{3}{*}{GCN} & Train ACC & -&-&-& 1.00 &0.84& 0.95 &-\\
& Valid ACC & -&-&-& 1.00 &1.00 & 0.86 &-\\
& Test ACC & -&-&-& 1.00 & 0.95 & 0.82 & -\\
\midrule
\multirow{3}{*}{GIN} & Train ACC & 0.99 & 1.00 & 0.97 &  1.00 &-& 0.93 &0.93 \\
& Valid ACC & 1.00 & 0.93 & 0.99 &  1.00 &-& 0.87 &0.87\\
& Test ACC & 0.97 & 0.95 & 0.97 &  1.00 &-& 0.89 &0.83\\
\bottomrule
\end{tabular}
\end{sc}
\end{center}
\end{table*}

\begin{table*}[ht]
\caption{Detailed qualitative results on \textbf{BA-2Motifs}. IG refers to Integrated Gradients. }
\vskip 0.1in
\label{tab:quality_detail_ba2motifs}
\begin{center}
\begin{sc}
\begin{tabular}{l|cc|cc}
\toprule
Method & (GCN) Class 0 &  (GCN) Class 1 & (GIN) Class 0 &  (GIN) Class 1 \\
\midrule
{\smodel}   & \includegraphics[width=0.135\textwidth]{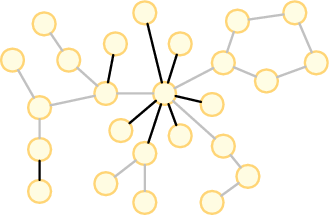} &\includegraphics[width=0.135\textwidth]{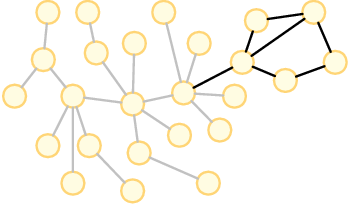} & \includegraphics[width=0.135\textwidth]{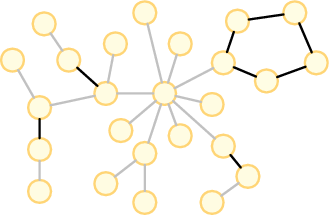} & \includegraphics[width=0.135\textwidth]{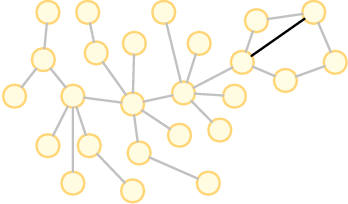}\\
\midrule
SA & \includegraphics[width=0.135\textwidth]{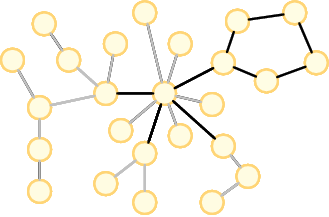} &\includegraphics[width=0.135\textwidth]{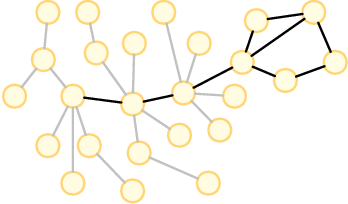}& \includegraphics[width=0.135\textwidth]{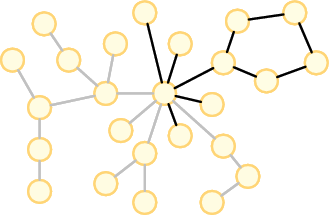}& \includegraphics[width=0.135\textwidth]{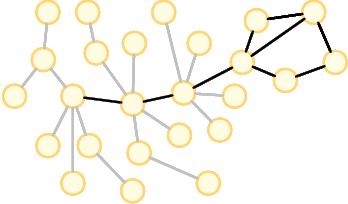}\\
\midrule
Grad-CAM & \includegraphics[width=0.135\textwidth]{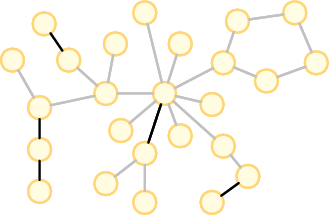} &\includegraphics[width=0.135\textwidth]{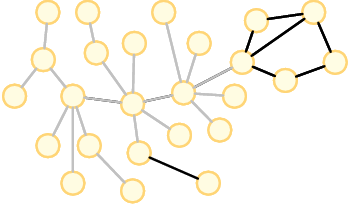}& \includegraphics[width=0.135\textwidth]{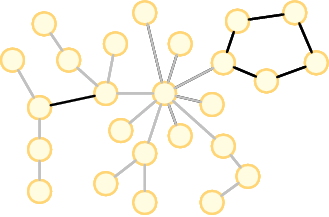}& \includegraphics[width=0.135\textwidth]{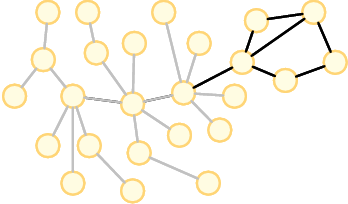}\\
\midrule
IG (Edge) & \includegraphics[width=0.135\textwidth]{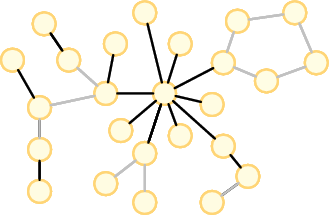} &\includegraphics[width=0.135\textwidth]{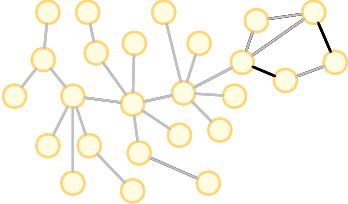}& \includegraphics[width=0.135\textwidth]{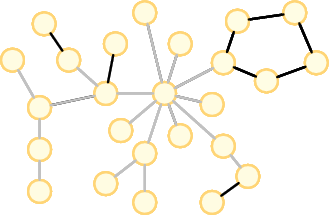}& \includegraphics[width=0.135\textwidth]{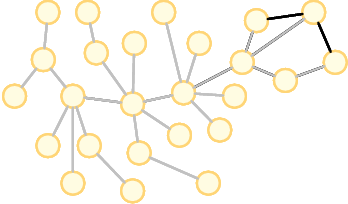}\\
\midrule
GNNExplainer & \includegraphics[width=0.135\textwidth]{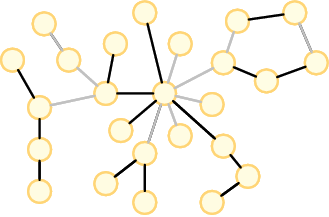} &\includegraphics[width=0.135\textwidth]{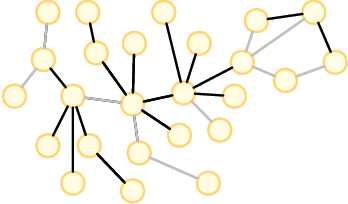}& \includegraphics[width=0.135\textwidth]{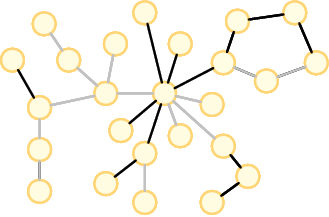}&\includegraphics[width=0.135\textwidth]{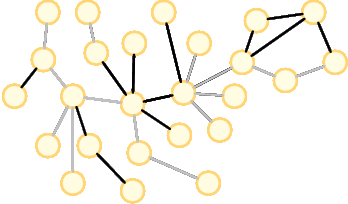}\\
\midrule
PGExplainer & \includegraphics[width=0.135\textwidth]{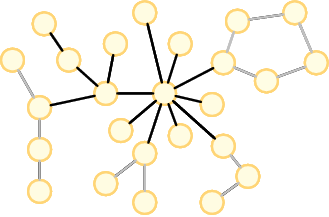} &\includegraphics[width=0.135\textwidth]{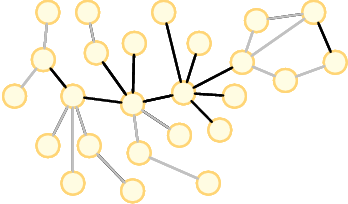}& \includegraphics[width=0.135\textwidth]{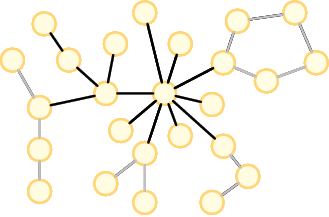}& \includegraphics[width=0.135\textwidth]{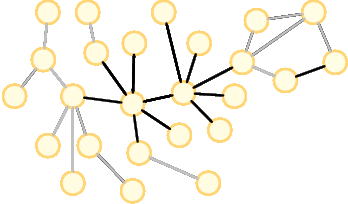}\\
\midrule
PGM-Explainer & \includegraphics[width=0.135\textwidth]{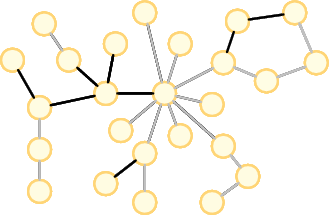} &\includegraphics[width=0.135\textwidth]{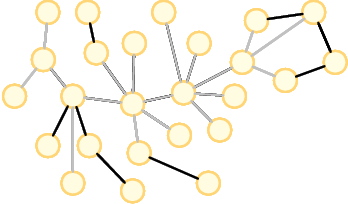}& \includegraphics[width=0.135\textwidth]{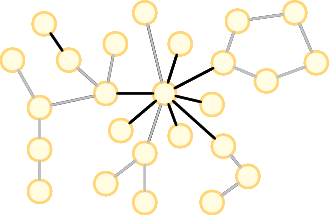}& \includegraphics[width=0.135\textwidth]{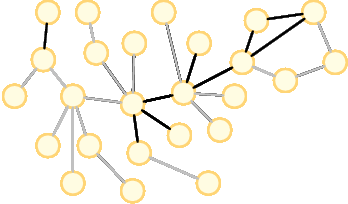}\\
\midrule
RCExplainer & \includegraphics[width=0.135\textwidth]{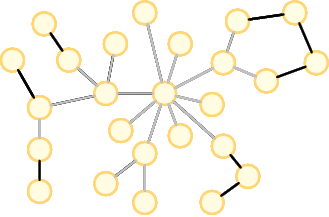} &\includegraphics[width=0.135\textwidth]{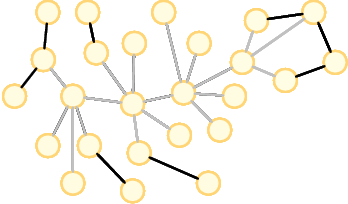}& \includegraphics[width=0.135\textwidth]{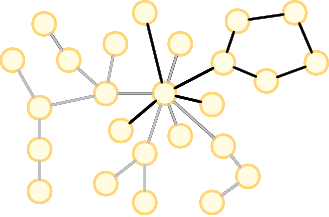}& \includegraphics[width=0.135\textwidth]{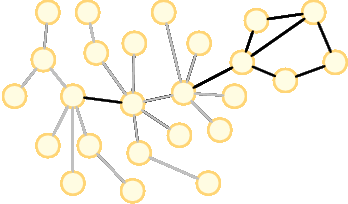}\\
\midrule
RG-Explainer & \includegraphics[width=0.135\textwidth]{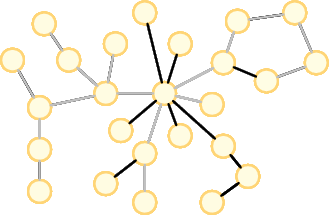} &\includegraphics[width=0.135\textwidth]{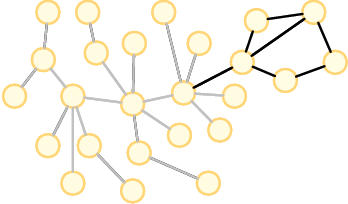}& \includegraphics[width=0.135\textwidth]{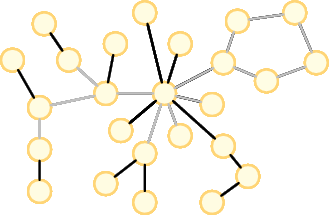}& \includegraphics[width=0.135\textwidth]{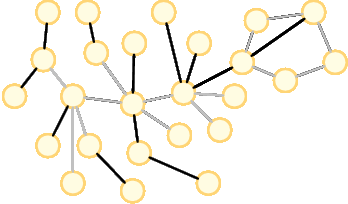}\\
\midrule
SubgraphX & \includegraphics[width=0.135\textwidth]{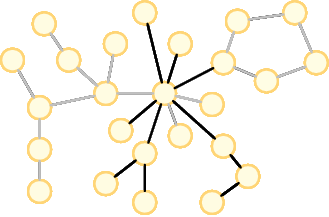} &\includegraphics[width=0.135\textwidth]{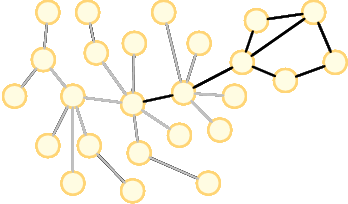}& \includegraphics[width=0.135\textwidth]{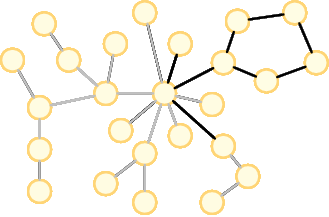}& 
\includegraphics[width=0.135\textwidth]{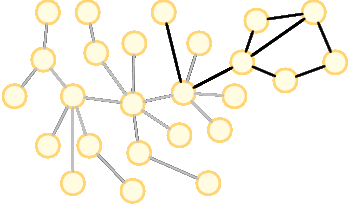}\\
\midrule
DEGREE & \includegraphics[width=0.135\textwidth]{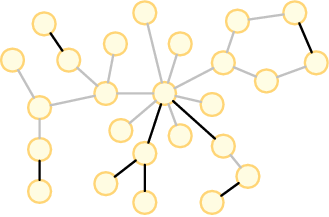} &\includegraphics[width=0.135\textwidth]{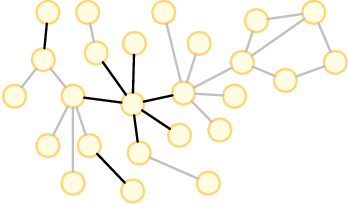}& \includegraphics[width=0.135\textwidth]{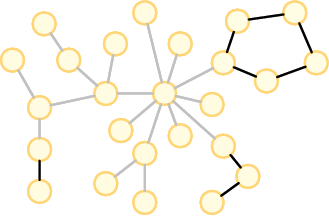}& \includegraphics[width=0.135\textwidth]{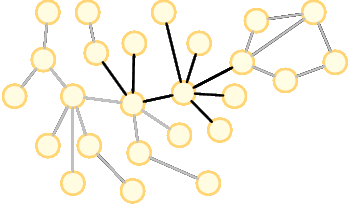}\\
\bottomrule
\end{tabular}
\end{sc}
\end{center}
\end{table*}

\begin{table*}[ht]
\caption{Detailed qualitative results on \textbf{Mutagenicity} and \textbf{NCI1}. IG refers to Integrated Gradients. }
\vskip 0.1in
\label{tab:quality_detail_other}
\begin{center}
\begin{sc}
\begin{tabular}{l|cc|c}
\toprule
Method & Mutagenicity (GCN) &  Mutagenicity (GIN) & NCI1 (GIN) \\
\midrule
{\smodel}   &\includegraphics[width=0.18\textwidth]{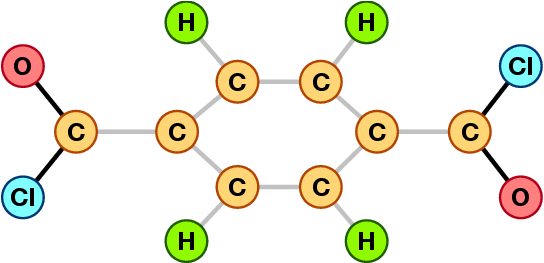} & \includegraphics[width=0.18\textwidth]{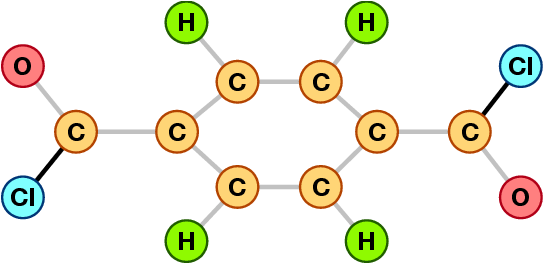} & \includegraphics[width=0.234\textwidth]{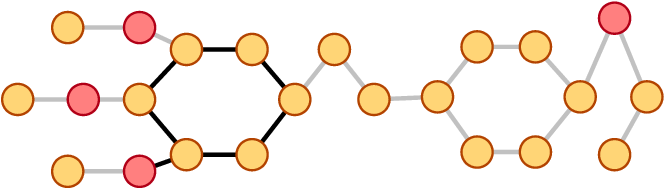}
\\
\midrule
SA &\includegraphics[width=0.18\textwidth]{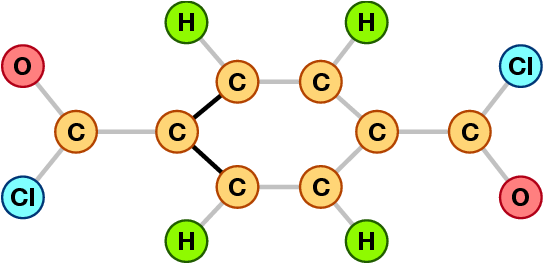}& \includegraphics[width=0.18\textwidth]{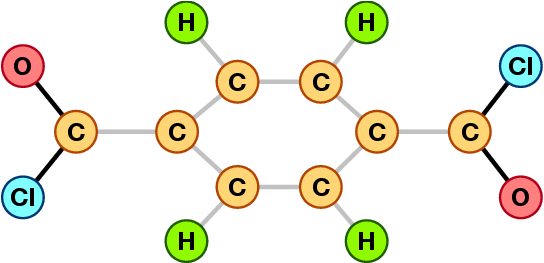}& \includegraphics[width=0.234\textwidth]{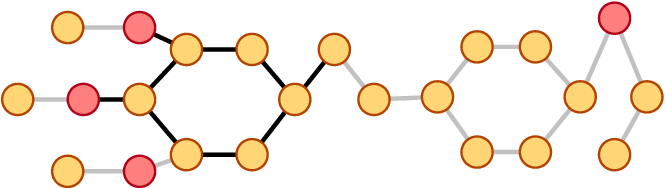}\\
\midrule
Grad-CAM &\includegraphics[width=0.18\textwidth]{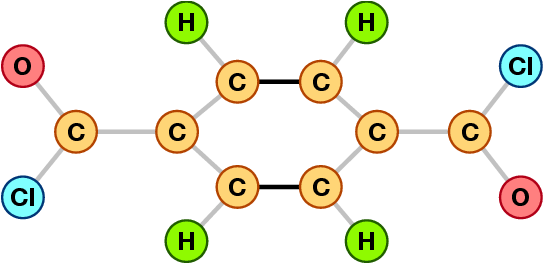}& \includegraphics[width=0.18\textwidth]{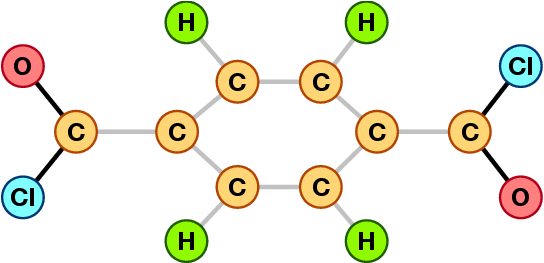}& \includegraphics[width=0.234\textwidth]{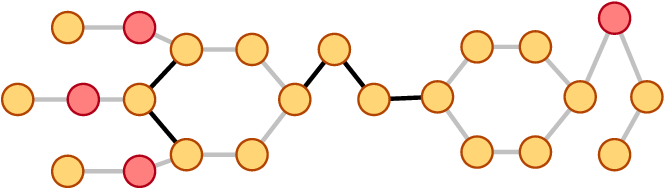}\\
\midrule
IG (Edge) &\includegraphics[width=0.18\textwidth]{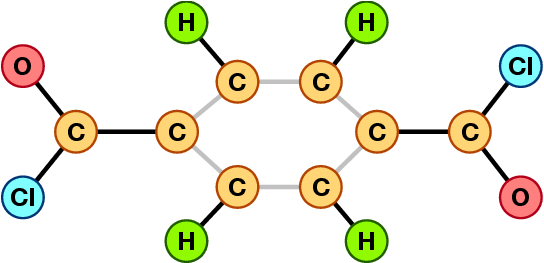}& \includegraphics[width=0.18\textwidth]{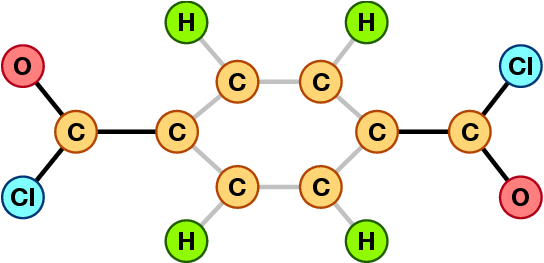}& \includegraphics[width=0.234\textwidth]{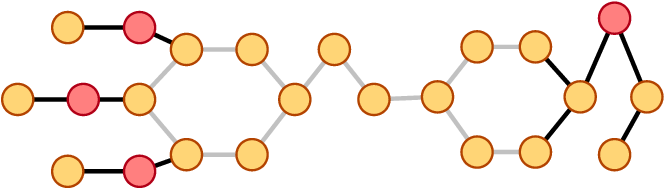}\\
\midrule
GNNExplainer &\includegraphics[width=0.18\textwidth]{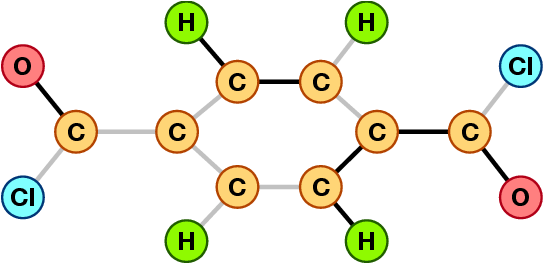}& \includegraphics[width=0.18\textwidth]{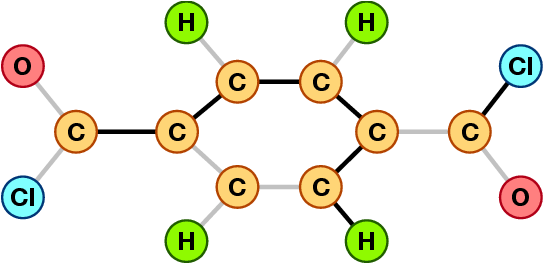}&\includegraphics[width=0.234\textwidth]{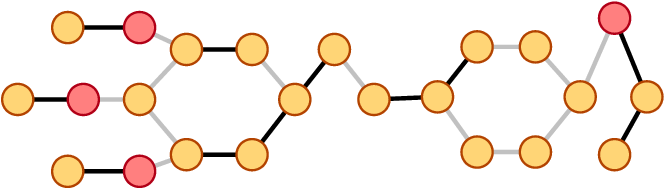}\\
\midrule
PGExplainer &\includegraphics[width=0.18\textwidth]{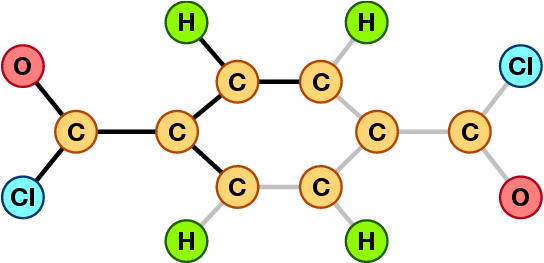}& \includegraphics[width=0.18\textwidth]{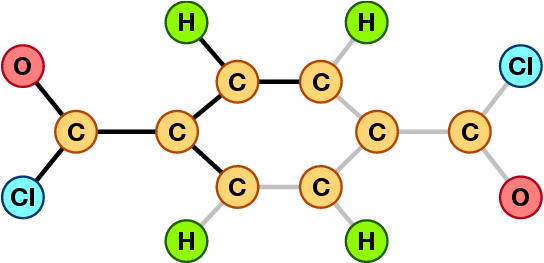}& \includegraphics[width=0.234\textwidth]{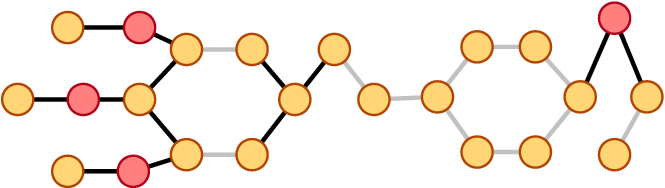}\\
\midrule
PGM-Explainer &\includegraphics[width=0.18\textwidth]{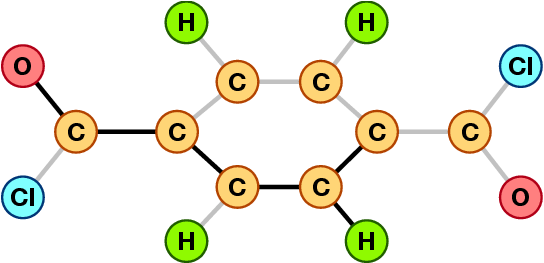}& \includegraphics[width=0.18\textwidth]{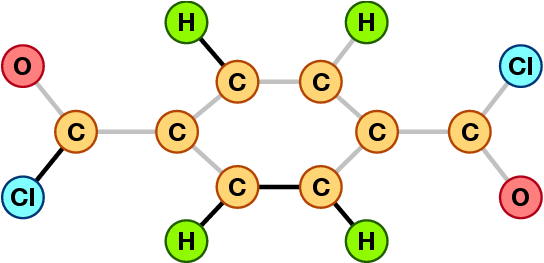}& \includegraphics[width=0.234\textwidth]{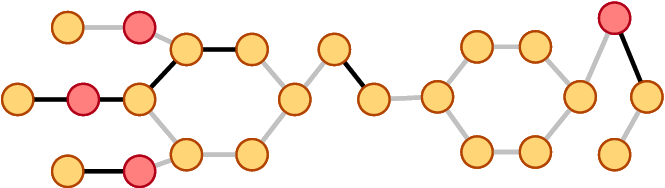}\\
\midrule
RCExplainer &\includegraphics[width=0.18\textwidth]{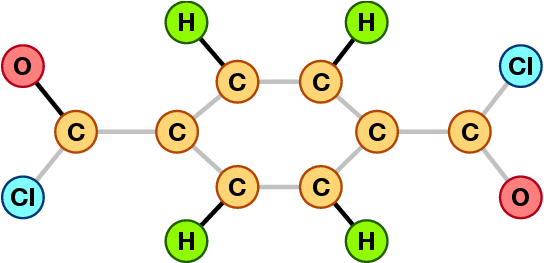}& \includegraphics[width=0.18\textwidth]{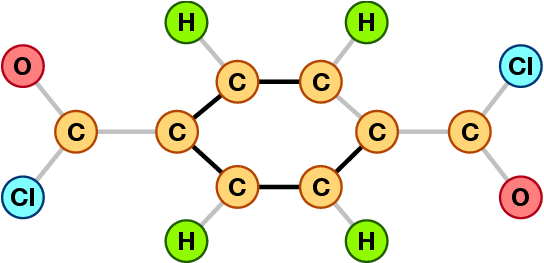}& \includegraphics[width=0.234\textwidth]{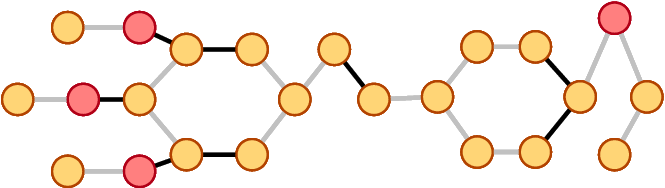}\\
\midrule
RG-Explainer &\includegraphics[width=0.18\textwidth]{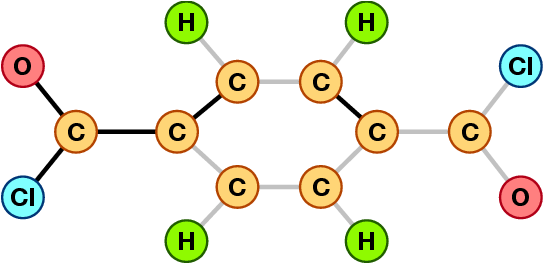}& \includegraphics[width=0.18\textwidth]{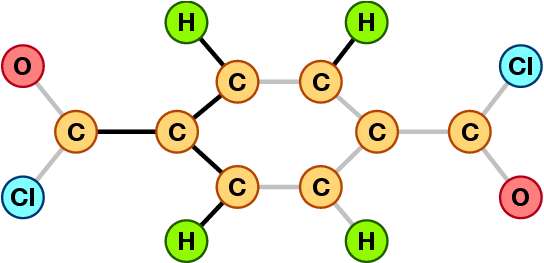}& \includegraphics[width=0.234\textwidth]{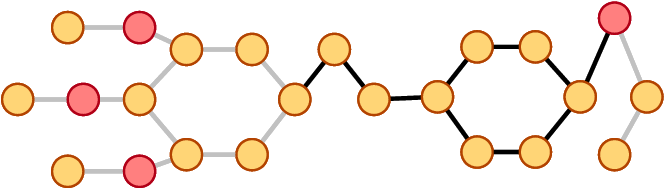}\\
\midrule
SubgraphX &\includegraphics[width=0.18\textwidth]{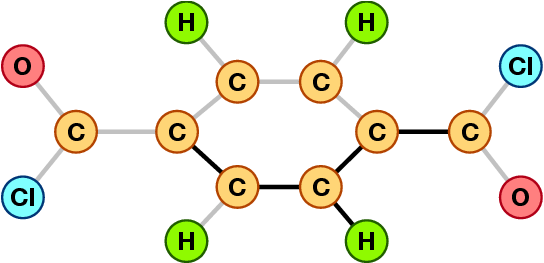}& \includegraphics[width=0.18\textwidth]{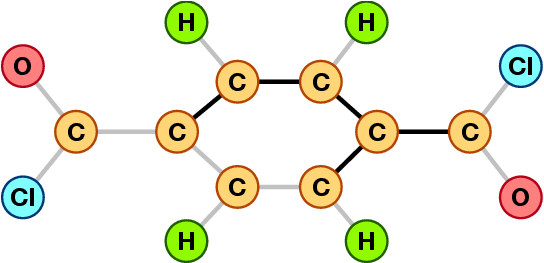}& 
\includegraphics[width=0.234\textwidth]{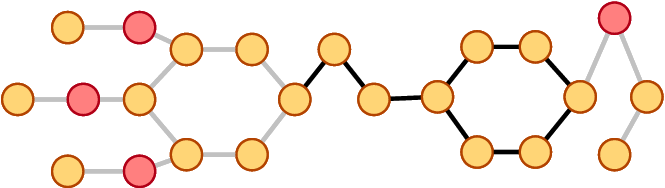}\\
\midrule
DEGREE &\includegraphics[width=0.18\textwidth]{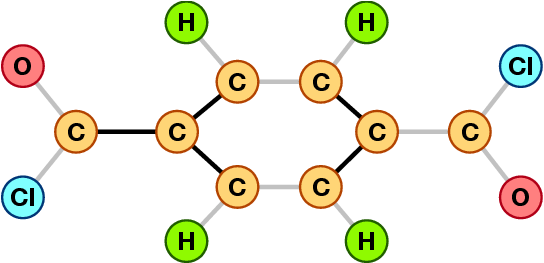}& \includegraphics[width=0.18\textwidth]{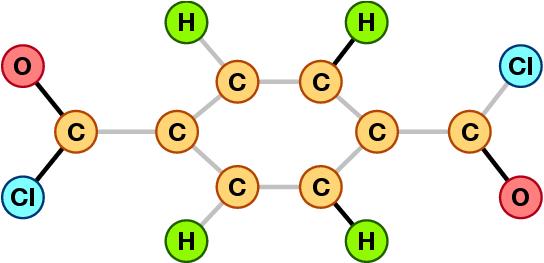}& \includegraphics[width=0.234\textwidth]{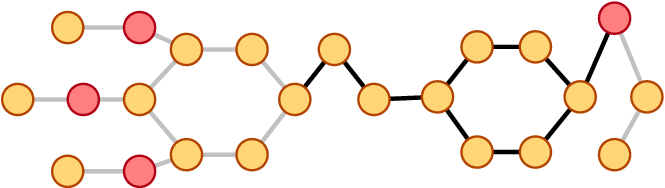}\\
\bottomrule
\end{tabular}
\end{sc}
\end{center}
\end{table*}


\end{document}